\documentclass[letterpaper, 10 pt, conference]{ieeeconf}  
\usepackage{amsmath}
\usepackage{graphicx} 
\usepackage[linesnumbered,ruled,vlined]{algorithm2e}
\usepackage{booktabs}  
\usepackage{adjustbox}  
\usepackage{float}  
\usepackage{multirow}
\usepackage{amssymb}
\usepackage{fancyhdr}
\usepackage{url}
\usepackage[utf8]{inputenc}
\usepackage{array}
\usepackage[utf8]{inputenc}
\usepackage{adjustbox}   
\usepackage{tabularx}    
\usepackage{booktabs}    
\usepackage{ragged2e}    
\usepackage{hyperref}
\usepackage{tcolorbox}
\usepackage{xltabular}
\usepackage{booktabs}   
\usepackage{graphicx}   
\usepackage{xcolor}     
\usepackage{hyperref}   
\usepackage{xltabular}  
\usepackage{xcolor}
\usepackage{graphicx}
\usepackage{caption}
\usepackage{hyperref}
\hypersetup{colorlinks=true, linkcolor=black, urlcolor=blue, citecolor=blue}
\usepackage{booktabs, array, ragged2e, tabularx, xltabular, xcolor, graphicx, hyperref}

\newcolumntype{Y}{>{\RaggedRight\arraybackslash}X}
\newcommand{\correct}[1]{\textcolor{orange}{\textbf{#1}}}
\renewcommand{\arraystretch}{1.15}

\newenvironment{OneColumnXLT}{
  \par\clearpage\onecolumn    
}{
  \par\clearpage\twocolumn    
}

\IEEEoverridecommandlockouts                              

\title{\LARGE \bf
AutoDrive-QA: A Multiple-Choice Benchmark for Vision–Language Evaluation in Urban Autonomous Driving
}
\author{Boshra Khalili$^{1}$ and Andrew W.Smyth$^{2}$
\thanks{$^{1}$Boshra Khalili is Graduate Research Assistant in the Department of Civil Engineering and Engineering Mechanics, Columbia University, New York, NY 10027, USA  {\tt\small bk2898@columbia.edu}}%
\thanks{$^{2}$Andrew W. Smyth (corresponding author) is the Robert A. W. and Christine S. Carleton Professor of Civil Engineering and Engineering Mechanics and the Director of the Center for the Smart Streetscapes (CS3), Columbia University, New York, NY 10027, USA {\tt\small  aws16@columbia.edu}}%
}

\begin{document}

\maketitle
\thispagestyle{empty}
\pagestyle{empty}
\pagenumbering{roman}
\newpage
\begin{abstract}

Evaluating vision–language models (VLMs) in urban driving contexts remains challenging, as existing benchmarks rely on open-ended responses that are ambiguous, annotation-intensive, and inconsistent to score. This lack of standardized evaluation slows progress toward safe and reliable AI for urban mobility. We introduce AutoDrive-QA, the first benchmark that systematically converts open-ended driving QA datasets (DriveLM, NuScenes-QA, LingoQA) into structured multiple-choice questions (MCQs) with distractors grounded in five realistic error categories: Driving Domain Misconceptions, Logical Inconsistencies, Misinterpreted Sensor Inputs, Computational Oversights, and Question Ambiguity. This framework enables reproducible and interpretable evaluation of VLMs across perception, prediction, and planning tasks in complex urban scenes. Experiments show that fine-tuning LLaVA-1.5-7B improves accuracy by about six percentage points across tasks, GPT-4V achieves the strongest zero-shot performance with up to 69.8\% accuracy, and Qwen2-VL models also perform competitively, particularly in multi-view settings. Moreover, traditional metrics such as BLEU and CIDEr fail to distinguish strong from weak models. By providing an objective, domain-grounded evaluation protocol, AutoDrive-QA contributes to more transparent benchmarking of urban AI systems, supporting the development of safer and more trustworthy autonomous driving technologies for smart cities. We release all the
codes in \href{https://github.com/Boshrakh/AutoDrive-QA}{AutoDrive-QA GitHub Repository}

\end{abstract}

\section{INTRODUCTION}
Autonomous driving in urban environments demands robust perception, accurate prediction, and reliable planning in order to navigate complex and dynamic scenes \cite{chen2024,nie2023,zhang2024,khalili2024}. Vision–language models (VLMs) have emerged as a promising direction for integrating multimodal driving information, yet their evaluation remains a persistent challenge \cite{huang2024,chen2024eval}. Existing QA datasets such as DriveLM \cite{sima2024}, CODA-LM \cite{chen2024eval}, and LingoQA \cite{marcu2024} rely on open-ended formats with heterogeneous annotations, making scoring inconsistent and cross-model comparison unreliable \cite{wang2023drivemlm,corbiere2025}. While open-ended responses capture reasoning, they are hard to evaluate. In contrast, multiple-choice QA offers objective scoring, reproducibility, and lower annotation cost \cite{mao2023}—yet no driving-specific MCQ benchmark currently exists for perception, prediction, and planning tasks \cite{tian2024}.
To address these gaps, we propose AutoDrive-QA, the first benchmark that converts open-ended autonomous driving QA datasets into structured multiple-choice questions (MCQs). In MCQs, distractors are the incorrect answer options, and their quality determines whether questions are trivial or genuinely test model understanding \cite{tian2024}. Prior benchmarks rarely model typical driving errors, and naïve distractor generation often produces options that are too obvious or irrelevant. AutoDrive-QA overcomes this with an LLM-driven pipeline that generates five categories of realistic distractors—Driving Domain Misconceptions, Logical Inconsistencies, Misinterpreted Sensor Inputs, Computational Oversights, and Question Ambiguity—spanning perception, prediction, and planning to approximate common failure modes in driving.

AutoDrive-QA complements open-ended QA by offering objective, reproducible evaluation and standardized baselines, while open-ended and human-in-the-loop testing remain essential for capturing reasoning depth and safety-critical reliability. Its systematic distractor generation and filtering mitigate trivial or duplicated options, grounding evaluation in realistic errors. By framing AutoDrive-QA as a baseline rather than a replacement, we aim to provide a reproducible, interpretable resource that enables fairer comparison across vision–language models for autonomous driving. This paper makes the following contributions:

\begin{itemize}

\item Benchmark creation. We introduce AutoDrive-QA, the first benchmark that converts open-ended autonomous driving QA datasets into structured multiple-choice questions (MCQs), enabling more standardized and reproducible evaluation of vision-language models (VLMs).
\item Distractor generation pipeline. We design an LLM-based distractor generation pipeline that systematically produces distractors across five categories of realistic driving errors Driving Domain including Misconceptions, Logical Inconsistencies, Misinterpreted Sensor Inputs, Computational Oversights, and Question Ambiguity across perception, prediction, and planning tasks, moving beyond generic or trivial options fosters genuine model understanding over guesswork, advancing the state-of-the-art in autonomous driving evaluation.
\item systematic error analysis. Using generated MCQs, we track which distractor a model selects when it makes an error, in order to determine whether its mistakes are visual, conceptual, or reasoning-based. This analysis reveals the strengths and weaknesses of VLM models in urban scene understanding and offers one of the direct comparisons of the types of errors—conceptual, reasoning, and visual—that VLM models exhibit across perception, prediction, and planning tasks. 

\end{itemize}

By addressing evaluation inconsistencies and domain-specific challenges, our research establishes a benchmark for autonomous driving VLMs. The paper is structured as follows: Section 2 reviews related work; Section 3 details the MCQ conversion methodology; Section 4 describes our experimental setup and analyzes the results; and Section 5 discusses implications for current models and future research directions.

\section{Related work}

Evaluating vision–language models for autonomous driving remains difficult due to the lack of standardized metrics and reliance on dataset-specific evaluation schemes \cite{fu2024}. Traditional measures such as BLEU \cite{papineni2002}, METEOR \cite{banerjee2005}, ROUGE \cite{lin2004}, and CIDEr \cite{vedantam2015} often perform poorly in driving-related QA, since responses may be correct even with low word overlap, or incorrect despite high n-gram similarity \cite{marcu2024}. GPT-based evaluations capture semantics more effectively but tend to overvalue fluent yet factually wrong answers\cite{marcu2024}. Further complicating matters, benchmarks like DriveLM \cite{sima2024}, LingoQA \cite{marcu2024}, MAPLM \cite{cao2024}, NuScenes-QA \cite{qian2024}, OmniDrive \cite{wang2024}, and NuInstruct \cite{ding2024} each adopt distinct rule-based, classifier-driven, or GPT-based procedures, making cross-dataset comparisons inconsistent \cite{fu2024}.

In this work, we propose AutoDrive-QA as a solution to these limitations. By converting open-ended QA into multiple-choice format with structured distractors, AutoDrive-QA enables objective, reproducible scoring while capturing domain-specific reasoning errors. A detailed discussion of evaluation metrics and a comparative analysis of existing datasets can be found in Appendix A and Appendix B, respectively. At the same time, improving evaluation is closely tied to how multiple-choice questions (MCQs) and distractors are generated, since the quality of distractors directly impacts whether benchmarks can measure genuine reasoning or merely test surface-level elimination strategies. This motivates a closer look at prior approaches to MCQ generation and distractor modeling \cite{zhang2025}.

\subsection{MCQ Generation and Distractor Modeling}

Traditional Multiple-Choice Question (MCQ) generators often rely on knowledge graphs \cite{yu2024}, reinforcement learning \cite{lu2022}, or retrieval techniques \cite{luo2024} to create distractors. Although these methods can produce plausible alternatives, the distractors are often too simple, off-topic, or require substantial human refinement \cite{zhang2025}. More recently, AutoConverter \cite{zhang2025}, a multi-agent LLM framework, has improved this process by proposing, reviewing, and refining distractors to ensure a single correct answer and sufficiently challenging alternatives, thereby reducing human effort and improving reproducibility. However, when applied to autonomous driving, AutoConverter falls short because it does not explicitly capture the common failure modes that arise in perception, prediction, and planning tasks.

To address this limitation, our approach tailors MCQ generation to the driving domain by introducing domain-specific prompts that produce distractors grounded in realistic misjudgments. Specifically, we classify common errors into five categories—Driving Domain Misconceptions, Logical Inconsistencies, Misinterpreted Sensor Inputs, Computational Oversights, and Question Ambiguity—and use this taxonomy to guide distractor construction. This ensures that distractors closely mirror real-world mistakes rather than trivial or generic errors. As a result, our method not only strengthens the assessment of model understanding in autonomous driving but also establishes a principled standard for constructing standardized evaluation frameworks in this field.

Building on these insights, we present AutoDrive-QA, the first benchmark to systematically convert driving QA datasets into multiple-choice format with structured, domain-grounded distractors. By aligning with established practices in general VLM evaluation while addressing critical gaps in driving benchmarks, AutoDrive-QA complements open-ended QA with a reproducible, standardized framework—representing a step toward more consistent, fair, and meaningful evaluation of vision–language models in autonomous driving.

\section{Method}

\subsection{Overview of AutoDrive-QA Pipeline}

\begin{figure*}
    \centering
    \includegraphics[width=0.9\linewidth]{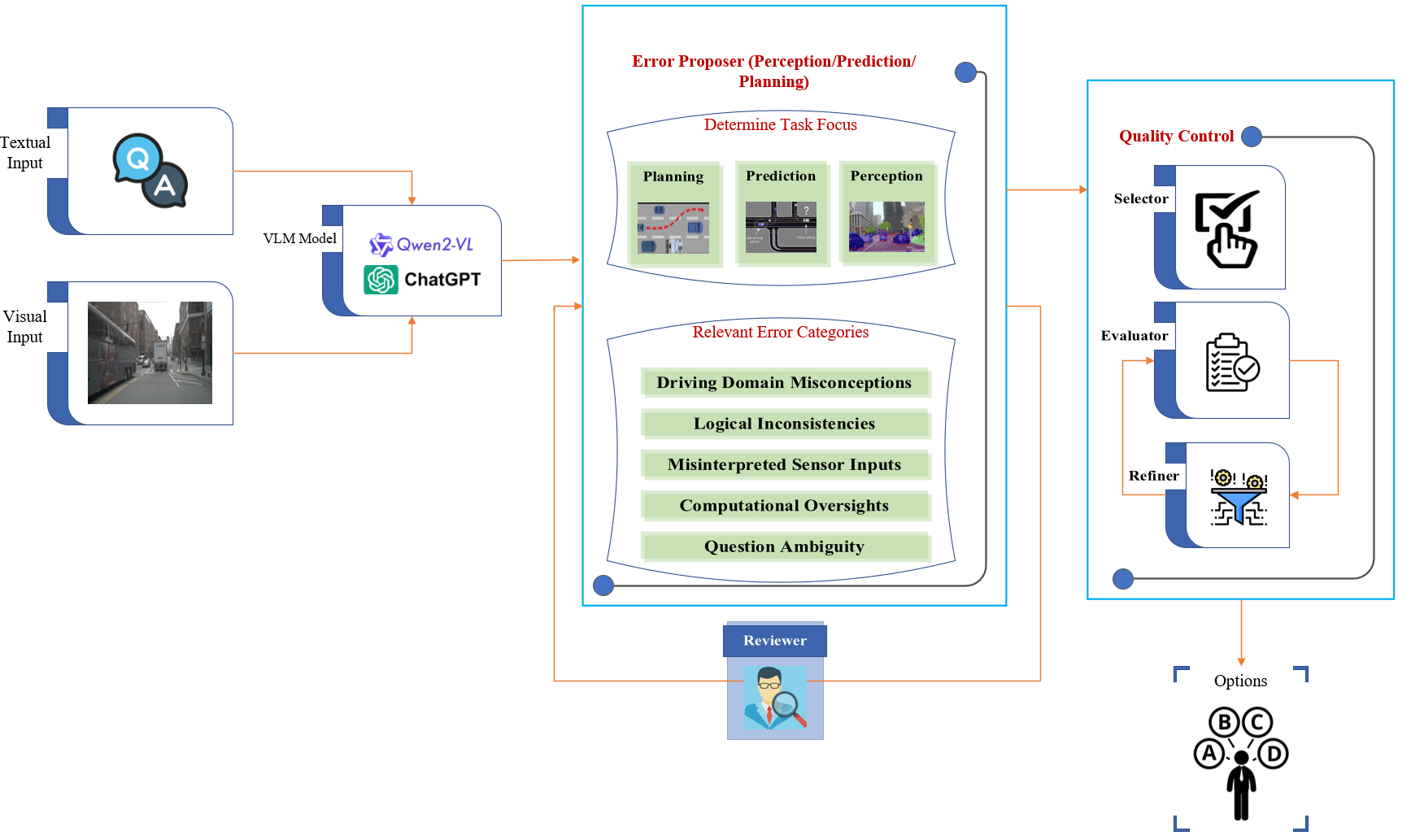}
    \caption{The AutoDrive-QA pipeline for automated MCQ generation in autonomous driving}
    \label{fig:pipline}
\end{figure*}

AutoDrive-QA introduces a systematic pipeline that transforms open-ended autonomous driving QA datasets into standardized multiple-choice questions (MCQs). Built on benchmarks such as DriveLM, NuScenes-QA, and LingoQA, the pipeline ensures that each question has a single correct answer, accompanied by distractors grounded in five error categories: driving domain misconceptions, logical inconsistencies, misinterpreted sensor inputs, computational oversights, and question ambiguity. These categories are designed to reflect common failure modes in autonomous driving. As illustrated in Figure~\ref{fig:pipline}, a multi-stage quality control process—consisting of  reviewer, evaluator, refiner, and selector stages—removes distractors that are irrelevant, redundant, or ambiguous, while randomizing answer order to avoid positional bias. To our knowledge, AutoDrive-QA is the first benchmark to adopt this structured approach in driving, providing an evaluation framework that eliminates ambiguity in open-ended scoring and enables consistent, reproducible comparison across vision–language models.

\subsection{Dataset Conversion}

To build AutoDrive-QA, we use existing autonomous driving QA datasets—DriveLM, NuScenes-QA, and LingoQA—which cover perception, prediction, and planning tasks. These datasets were originally open-ended, meaning models produced free-form answers. While flexible, such answers are difficult to evaluate because they can be interpreted in different ways and often depend on subjective judgment. AutoDrive-QA addresses this by converting each question into a multiple-choice format (MCQ) with one correct answer from the dataset and additional distractors generated by our LLM-based pipeline. This ensures that each question has exactly one correct answer, making evaluation objective, consistent, and scalable. Although this simplifies some reasoning, it removes scoring ambiguity, enables fair model comparison, and supports systematic error analysis. Explanation and representative examples of the converted dataset are provided in Appendix C.

\subsection{Distractor Generation Pipeline}

A core contribution of AutoDrive-QA is its structured distractor generation pipeline, which converts open-ended QA items into robust multiple-choice format. Unlike benchmarks with trivial distractors, our pipeline encodes autonomous driving knowledge so that each distractor reflects plausible yet flawed reasoning, preventing answers from being guessed by elimination and making each MCQ a genuine test of understanding.
Autonomous driving tasks can be broadly grouped into three categories: Perception, which involves surrounding object identification, traffic element identification, important object identification, visual descriptions, and motion states; Prediction, which includes motion prediction, possible attention, object interaction, logical sequence, signal meaning, and visual occlusion; and Planning, which encompasses safe/unsafe action, goal action, possible collision, free-form QA/comment, importance ranking, planning and reasoning, and ego attention.
To ensure domain specificity, candidate distractors are generated by LLMs but guided by five predefined error categories. These include: Driving Domain Misconceptions, which capture fundamental misunderstandings of road rules or object categories; Logical Inconsistencies, which reflect flawed reasoning or incorrect cause-and-effect assumptions; Misinterpreted Sensor Inputs, representing errors in processing visual or sensor data; Computational Oversights, covering mistakes in quantitative reasoning such as misjudging speed, distance, or collision time; and Question Ambiguity, which arises from unclear or complex phrasing.

By grounding distractors in these real-world failure modes, AutoDrive-QA ensures that they are both plausible and clearly flawed. This approach not only reduces guesswork but also enables a targeted assessment of perception, prediction, and planning skills in autonomous driving systems. The full specifications of the distractor categories and prompt designs are provided in Appendix H.

\subsection{Filtering and Validation}

After the initial distractor generation, we use a multi-stage quality control process to ensure the final MCQs are both challenging and unambiguous. Our framework employs a multi-agent process to ensure both difficulty and correctness of multiple-choice questions. Proposers first generate candidate distractors based on common error types. A reviewer evaluates them and provides feedback, which the refiner uses to improve their plausibility while keeping them incorrect. The selector then chooses the three most challenging distractors, and the evaluator verifies that exactly one correct answer remains, triggering refinement if ambiguity is detected. The full specifications of the distractor categories and prompt designs are provided in Appendix I.

\section{Experiments}

\subsection{Dataset Construction and Statistics}
The AutoDrive-QA benchmark is built by converting questions from three open-ended autonomous driving QA datasets—DriveLM, NuScenes-QA, and LingoQA—into a standardized multiple-choice format, spanning perception, prediction, and planning tasks. Specifically, 60\% of questions are drawn from DriveLM, 20\% from NuScenes-QA, and 20\% from LingoQA. Each open-ended QA instance is reformulated into one question with a single correct answer and three distractors derived from our structured error categories, ensuring that distractors reflect realistic mistakes rather than trivial options. 

The benchmark contains 15{,}400 curated questions, balanced across categories: perception (40\%), prediction (30\%), and planning (30\%), with each category including at least 4{,}000 questions to enable robust evaluation. To maintain quality, we applied strict filtering so that no distractor duplicated the correct answer. A three-stage review process (\textit{Reviewer $\rightarrow$ Evaluator $\rightarrow$ Refiner}) removed irrelevant, redundant, or inadvertently correct distractors, and the option order was randomized to avoid positional bias. To validate clarity and answerability, human experts solved 400 randomly sampled questions. They achieved nearly 100\% accuracy on perception and over 97.5\% accuracy on prediction and planning, confirming that the questions are unambiguous for domain experts while still challenging for VLMs. The most common human errors involved overlooking hazards, misjudging following speed, and misclassifying parked cars. A more detailed description of dataset construction, filtering stages, and representative examples is 
provided in Appendix C.

\section{Evaluation of SOTA Vision--Language Models}

Table~\ref{tab:Evaluating_vision-language} reports the performance of several vision--language models on the AutoDrive benchmark. Fine-tuning LLaVA-1.5-7B yields consistent improvements, increasing perception accuracy from 57.84\% to 63.16\%, prediction from 52.70\% to 58.84\%, and planning from 53.66\% to 58.15\%. These results correspond to gains of approximately 6 percentage points across all three tasks, underscoring the benefit of task-specific adaptation.  

Among zero-shot models, GPT-4V attains the highest accuracies, with 69.81\% in perception, 66.25\% in prediction, and 68.5\% in planning. Qwen2-VL-72B follows with strong results of 66.2\%, 64\%, and 64.84\%, while Qwen2-VL-7B also achieves competitive performance across tasks.  
The results further highlight the benefits of multi-view reasoning: Qwen2-VL-7B achieves higher accuracies than its single-view counterpart, improving perception from 60.1\% to 63.54\%, prediction from 58.05\% to 60.64\%, and planning from 59\% to 61.02\%. Finally, BLEU and CIDEr scores remain relatively low across all models (ranging from 5.9--8.5 for BLEU and 27.1--39.2 for CIDEr), reinforcing the observation that traditional n-gram--based metrics are poorly aligned with task success, whereas AutoDrive accuracy provides a more reliable measure of model performance.

\begin{table*}[t]
\centering
\caption{Evaluation of vision--language models on the AutoDrive benchmark. Performance is reported for Perception, Prediction, and Planning. Fine-tuning LLaVA-1.5-7B yields 5--6 percentage point improvements across tasks, while GPT-4V achieves the highest zero-shot accuracy. Qwen2-VL models also perform competitively, with multi-view reasoning providing additional gains. BLEU and CIDEr remain consistently low across all models.}

\label{tab:Evaluating_vision-language}
\resizebox{\textwidth}{!}{%
\begin{tabular}{@{}ll*{9}{c}@{}}
\toprule
\textbf{Category} & \textbf{No. Views} 
& \multicolumn{3}{c}{\textbf{Perception}} 
& \multicolumn{3}{c}{\textbf{Prediction}} 
& \multicolumn{3}{c}{\textbf{Planning}} \\ 
\cmidrule(lr){3-5}\cmidrule(lr){6-8}\cmidrule(lr){9-11}
 & & \textbf{AutoDrive (\%, ↑)} & \textbf{BLEU(↑)} & \textbf{CIDEr(↑)} 
   & \textbf{AutoDrive (\%, ↑)} & \textbf{BLEU(↑)} & \textbf{CIDEr(↑)} 
   & \textbf{AutoDrive (\%, ↑)} & \textbf{BLEU(↑)} & \textbf{CIDEr(↑)}  \\ 
\midrule
LLaVA-1.5-7B \textit{(fine-tuned)} & 1 
& 63.16  & 6.5  & 29.6
& 58.84  & 6.92  & 31.4
& 58.15  & 6.88  & 28.5\\
LLaVA-1.5-7B \textit{(zero-shot)} & 1 
& 57.84 & 5.9 & 28.1  
& 52.7 & 5.25 & 30.2 
& 53.66 & 5.3 & 27.1 \\
\midrule
GPT-4V \textit{(zero-shot)} & $\geq 1$ 
& 69.81 & 7.7 &  35.11
& 66.25 & 5.5 & 32.5 
& 68.5 & 5.25 & 31.65 \\
Qwen2-VL-72B \textit{(zero-shot)} & $\geq 1$ 
& 66.2 & 8.5 & 39.2 
& 64 & 5.95 & 32.81 
& 64.84 & 6.89 & 35.69 \\
Qwen2-VL-7B \textit{(zero-shot)} & $\geq 1$ 
& 63.54 & 6.25 & 30.5 
& 60.64 & 6.22 &  28.9
& 61.02 & 5.9 & 29.66 \\
Qwen2-VL-7B \textit{(zero-shot)} & 1 
& 60.1 & 6.06 & 30.9 
& 58.05 & 6.1 &  29.5
& 59 & 5.95 & 29.25 \\
\bottomrule
\end{tabular}%
}
\end{table*}

\subsection{Additional Results and Analysis}

Beyond the main cross-model results, we further examine how models behave under different conditions. First, we provide qualitative examples (Appendix D) that illustrate how AutoDrive-QA captures subtle reasoning mistakes in urban driving. At the per-example level, traditional metrics such as BLEU and CIDEr often assign partial credit to incorrect answers, whereas AutoDrive accuracy aligns strictly with correctness. This highlights the limitations of n-gram-based metrics and underscores AutoDrive-QA’s value as a diagnostic, task-relevant evaluation framework.

Next, we analyze errors by distractor type (Appendix F) to determine whether models struggle more with visual details, conceptual understanding, or reasoning. Table~\ref{table:Error-Modeling-Performance} shows that perception and prediction errors are primarily driven by \emph{driving domain misconceptions} (45.7\% and 41.1\%, respectively), while planning errors are dominated by \emph{logical inconsistencies} (48.9\%). These results suggest that models continue to face critical challenges in reasoning about complex interactions. More detailed analyses of systematic error patterns are provided in Appendix F.

In addition, we conduct robustness checks to examine the impact of distractor design and filtering (Appendix G). Ablation experiments (Table~\ref{table:Error-Modeling}) reveal that accuracy drops by 7--9\% when realistic distractors are used in place of naïve alternatives. This confirms that AutoDrive-QA does not merely test superficial elimination strategies but instead requires genuine scene understanding and reasoning.

\section{Conclusion }

While open-ended formats provide valuable insights into generative reasoning and diverse phrasing, they remain difficult to grade consistently across models. Our structured MCQ framework addresses this challenge by introducing a standardized and reproducible evaluation protocol for urban driving scenarios. It is designed to complement open-ended QA while also laying the foundation for richer and more robust benchmarks in Urban AI and smart mobility systems. Based on our experimental results, fine-tuning LLaVA-1.5-7B improved accuracy by approximately six percentage points across perception, prediction, and planning tasks. GPT-4V delivered the strongest zero-shot performance, reaching up to 69.8\% accuracy. Qwen2-VL models also performed competitively—particularly in multi-view settings—while BLEU and CIDEr scores remained consistently low, underscoring the misalignment of traditional text-overlap metrics with driving task success.

Looking ahead, AutoDrive-QA opens several promising directions. These include expanding to more diverse datasets for broader city-scale coverage, adopting modular architectures that separate visual feature extraction from large language models, and advancing video understanding through domain-specific fine-tuning. Another important step is linking benchmark evaluation with urban traffic simulation environments, thereby connecting question-answering performance directly to downstream driving behavior. In addition, tailoring the framework to specialized models (e.g., DriveVLM) could further enhance its real-world applicability. In conclusion, AutoDrive-QA offers a complementary and evolving framework for evaluating vision–language models in autonomous urban driving, paving the way toward more reliable, safer, and human-centered AI-driven mobility systems for smart cities.


\appendix

\section*{Appendix A: Evaluation Metrics}
\addcontentsline{toc}{section}{Appendix A: Evaluation Metrics}
\label{Evaluation Metrics}
Evaluating open-ended responses for autonomous driving requires deep contextual understanding and rigorous reasoning. Traditional rule-based metrics such as BLEU \cite{papineni2002}, ROUGE \cite{lin2004}, METEOR \cite{banerjee2005}, CIDEr \cite{vedantam2015}, and SPICE \cite{anderson2016} provide quick assessments based on lexical overlap with reference answers. However, these metrics capture only surface-level similarity, overlooking semantic depth and multi-step logical reasoning that are critical in autonomous driving \cite{wen2023}. As a result, they correlate weakly with human judgments and fail to measure nuanced contextual understanding \cite{wang2023}.

To address these limitations, model-based approaches have been proposed. These methods employ classifiers or large language models with built-in reasoning and factual checks to approximate human evaluation \cite{jiang2023}. For example, Lingo-Judge \cite{marcu2024} leverages DeBERTa-V3 \cite{he2021}, while other approaches rely on GPT-4 \cite{openai2023}. Although such models better capture semantic meaning and reasoning, they come with drawbacks: Lingo-Judge risks overfitting to dataset-specific styles, and GPT-4’s results can vary with prompt design or model version updates, raising concerns about consistency and reproducibility \cite{chib2023}.

In this work, we introduce AutoDrive-QA, a novel MCQ-based evaluation framework for autonomous driving. While benchmarks such as VQA-v2, GQA, ScienceQA, MMMU, and MMBench have long adopted multiple-choice questions for objective scoring and fair cross-model comparison, autonomous driving evaluation has remained dominated by open-ended QA. AutoDrive-QA fills this gap by extending MCQ evaluation to driving-specific scenarios. Using carefully designed structured distractors, it converts ambiguous free-text answers into measurable reasoning errors, providing a standardized, reproducible, and domain-grounded evaluation method.
\section*{Appendix B: Evaluation metrics in Question Answering Datasets for Autonomous Driving}
\addcontentsline{toc}{section}{Appendix B: Evaluation metrics in Question Answering Datasets for Autonomous Driving}
\label{Evaluation metrics-QA}

Different datasets use a variety of evaluation metrics, which creates challenges for data integration and generalization. For example, DriveLM \cite{sima2024} combines real-world nuScenes \cite{caesar2020} data with simulated CARLA \cite{dosovitskiy2017} data to test multi-step reasoning in perception, prediction, and planning. Its evaluation framework employs rule-based metrics, completeness scores, trajectory errors, and GPT-based checks. In contrast, LingoQA \cite{marcu2024} captures driver rationales from short video clips by using rule-based metrics, a specialized classifier (Lingo-Judge), and GPT-4V evaluations. MAPLM \cite{cao2024} assesses HD map understanding by comparing exact or semantic matches to ground-truth answers, although it may overlook partially correct responses. NuScenes-QA \cite{qian2024} emphasizes 3D reasoning with exact or soft-match criteria, which can oversimplify complex answers. OmniDrive \cite{wang2024} builds on nuScenes by incorporating 3D and counterfactual queries evaluated by GPT-based or custom judges; however, it lacks a consistent universal metric. Finally, NuInstruct \cite{ding2024} presents 91K multi-view video-instruction pairs across perception, prediction, and planning, using GPT-based semantic evaluations that provide flexibility but complicate reproducibility. Each dataset targets a distinct aspect of autonomous driving and employs evaluation techniques ranging from basic lexical checks to advanced GPT-based assessments. However, balancing semantic detail, cost, and standardization remains challenging. Table~\ref{tabel:different_data} compares these datasets by detailing their scale, domain focus, input modalities, tasks, Q/A formats, and evaluation methods.
\begin{table*}
\caption{Comparison of Autonomous Driving Datasets}
\centering
{\fontsize{7}{8}\selectfont
\begin{tabularx}{\textwidth}{@{}l l >{\raggedright\arraybackslash}X@{}}
\toprule
\textbf{Dataset} & \textbf{Tasks Covered} & \textbf{Evaluation Methods} \\
\midrule
DriveLM\newline\cite{sima2024} 
& Object detection, trajectory prediction, planning 
& Rule-based; completeness; trajectory error; GPT-based checks \\
\midrule
LingoQA\newline\cite{marcu2024} 
& Scene description; action justifications (“why”/“how”) 
& Lingo-Judge classifier; rule-based; GPT-4 evaluations \\
\midrule
MAPLM (MAPLM-QA)\newline\cite{cao2024} 
& Vision–language QA on map elements 
& QA correctness (exact / semantic match) \\
\midrule
CODA-LM\newline\cite{chen2024} 
& Regional and general perception; driving suggestions 
& GPT-4 judge (1--10 scale); baseline n-gram metrics \\
\midrule
NuScenes-QA\newline\cite{qian2024} 
& Counting, attribute comparison, spatial relations 
& Exact / soft-match accuracy \\
\midrule
OmniDrive\newline\cite{wang2024} 
& Scene description; 3D grounding; counterfactual reasoning; planning 
& Specialized domain scoring (GPT-based or custom judges) \\
\midrule
NuInstruct\newline\cite{ding2024} 
& Perception, prediction, risk assessment, planning with reasoning 
& Semantic / human or GPT-based checks \\
\bottomrule
\end{tabularx}
}
\label{tabel:different_data}
\end{table*}

\section*{Appendix C: AutoDrive-QA Dataset Examples}
\addcontentsline{toc}{section}{Appendix C: Examples of AutoDrive-QA}
\label{Examples of AutoDrive-QA}
Each question–answer pair in our dataset is associated with a unique scene identifier and keyframe, thereby ensuring precise temporal and spatial grounding. As an illustrative example, consider the case presented in the last row of Table~\ref{tab:S_AutoDrive_examples}. The corresponding question is “Is <c2,CAM\_FRONT,708.7,571.4> an object that the ego vehicle should consider in the current scene?” with the annotated correct answer “Yes.” The token <c2,CAM\_FRONT,708.7,571.4> denotes a key object characterized by (i) a unique identifier (c2), (ii) the camera view from which it is observed (CAM\_FRONT), and (iii) its approximate two-dimensional position within the image plane. Each key object is further enriched with semantic and geometric attributes, including category (e.g., vehicle, traffic element), status (e.g., stationary, moving, null), visual description (e.g., “Blue bus,” “Traffic signal”), and bounding-box coordinates. In this specific example, the referenced object corresponds to a stationary blue bus captured by the front-facing camera. This structured representation establishes a direct link between questions, answers, and scene content, thereby guaranteeing grounding at both the scene and object levels. Moreover, it enables the integration of object-level annotations into subsequent processes such as distractor generation and evaluation. 
Table~\ref{tab:S_AutoDrive_examples} provides representative success cases from AutoDrive-QA, covering tasks across perception, prediction, and planning. Each entry illustrates the task type, the corresponding camera image, the natural-language question, and the multiple-choice options, with the correct answer highlighted in orange. In addition to success cases, Table~\ref{tab:AutoDrive_examples} presents representative examples of failure cases across perception, prediction, and planning. These examples demonstrate typical error modes, such as misclassifying static objects as dynamic, overlooking developing hazards, or recommending unsafe maneuvers, thereby illustrating how AutoDrive-QA captures subtle reasoning failures that are often missed by traditional evaluation metrics.

\begin{OneColumnXLT}  
\begin{center}
\begin{xltabular}{\textwidth}{@{}l c p{4cm} Y@{}}
\caption{Success cases of AutoDrive-QA. These are examples where the model makes a correct judgment about the correctness of the prediction. Green indicates agreement with ground truth, while red indicates disagreement. Each example lists the task, image, question, and four choices (with the correct choice shown in \textcolor{orange}{orange}).}
\label{tab:S_AutoDrive_examples}\\
\toprule
\textbf{Task} & \textbf{Image} & \textbf{Question} & \textbf{Choices} \\
\midrule
\endfirsthead

\multicolumn{4}{c}{{\tablename\ \thetable} -- continued from previous page}\\
\toprule
\textbf{Task} & \textbf{Image} & \textbf{Question} & \textbf{Choices} \\
\midrule
\endhead

\midrule
\multicolumn{4}{r}{Continued on next page}\\
\endfoot

\bottomrule
\endlastfoot

\hypertarget{ex1}{}%
Perception &
\includegraphics[width=1.8cm,height=1.8cm,keepaspectratio]{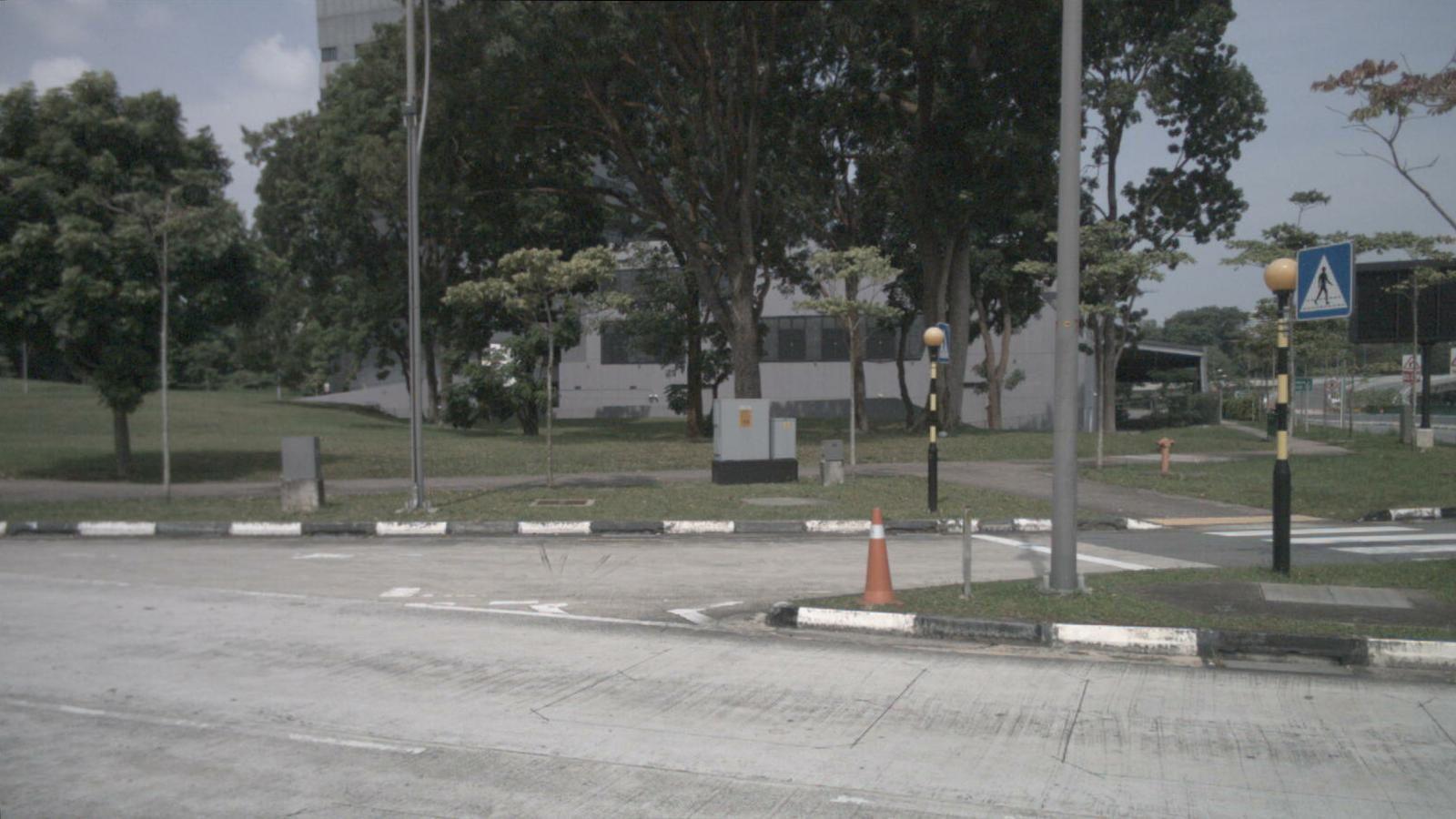} &
What are objects to the back left of the ego car? &
\correct{A. There is one traffic cone to the back left of the ego car.}\\
& & & B. There are multiple pedestrians crossing behind the ego car.\\
& & & C. A vehicle is partially visible behind the ego car on the left.\\
& & & D. A pedestrian is standing near the curb to the back left.\\[2pt]

\hypertarget{ex2}{}%
Perception &
\includegraphics[width=1.8cm,height=1.8cm,keepaspectratio]{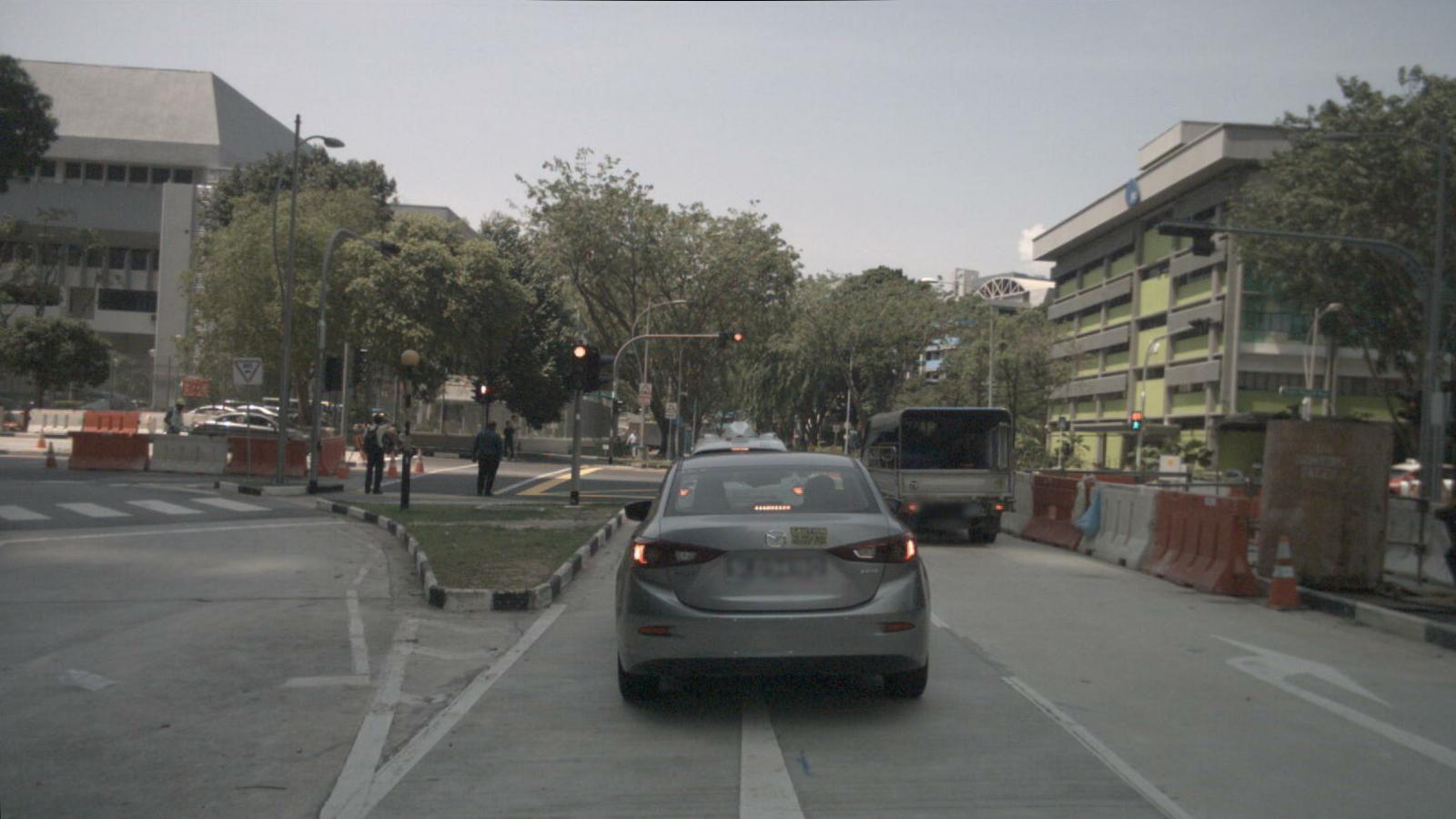} &
What are objects to the front of the ego car? &
A. There are no vehicles in front of the ego car because the traffic light is red.\\
& & & B. A cyclist is directly in front of the ego car.\\
& & & \correct{C. There are many barriers, one traffic cone, many pedestrians, one truck, and three cars in front of the ego car.}\\
& & & D. There can’t be any objects ahead.\\[2pt]

\hypertarget{ex3}{}%
Prediction &
\includegraphics[width=1.8cm,height=1.8cm,keepaspectratio]{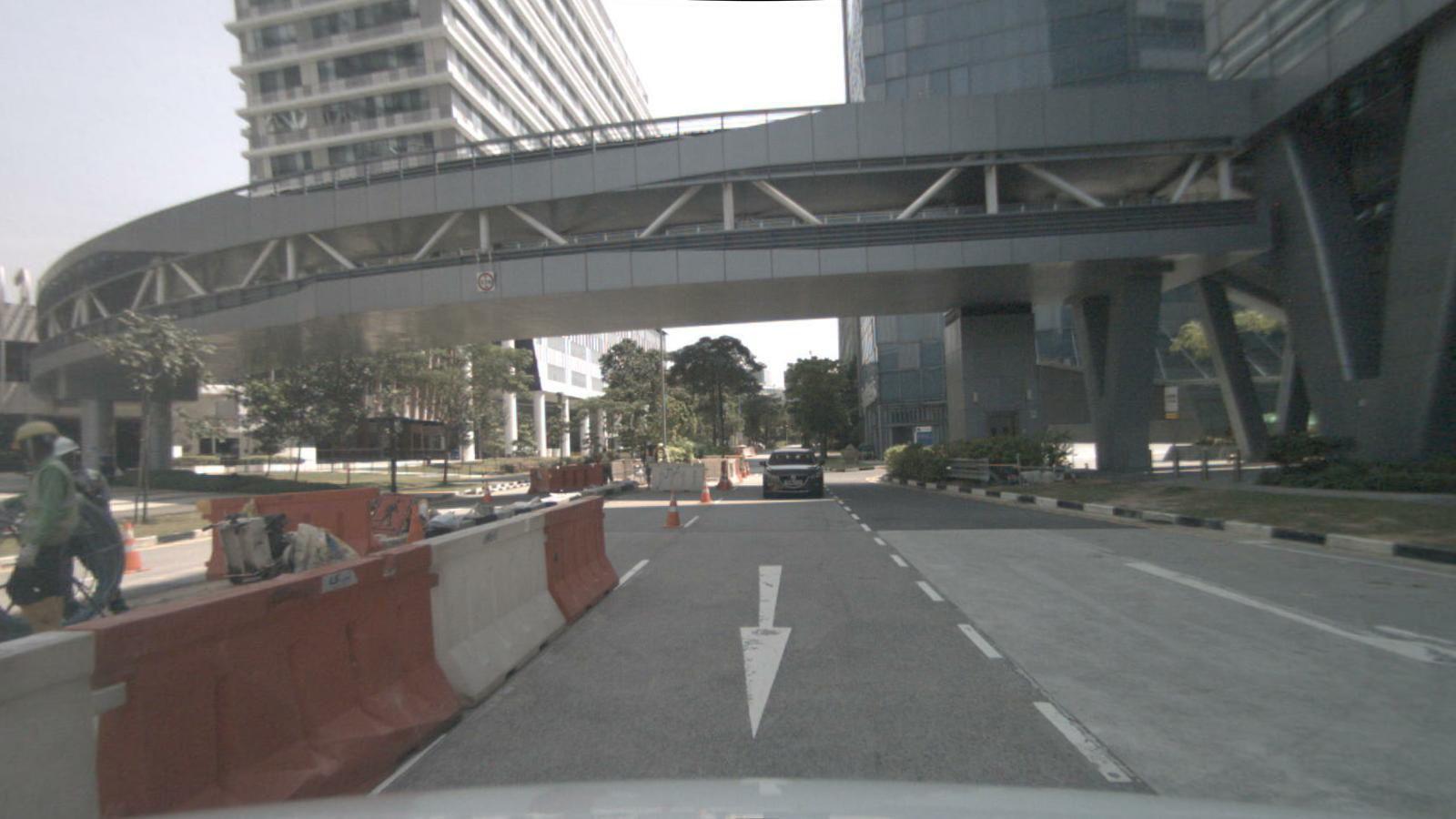} &
Except for the ego vehicle, what object would consider \texttt{<c2,\allowbreak CAM\_BACK,\allowbreak 874.2,\allowbreak 519.2>}
 to be most relevant to its decision? &
A. The following vehicle is always the most relevant object.\\
& & & B. A cyclist is at that location, moving toward the ego car.\\
& & & \correct{C. None.}\\
& & & D. The traffic sign on the bridge is a ``Stop'' sign.\\[2pt]

\hypertarget{ex4}{}%
Prediction &
\includegraphics[width=1.8cm,height=1.8cm,keepaspectratio]{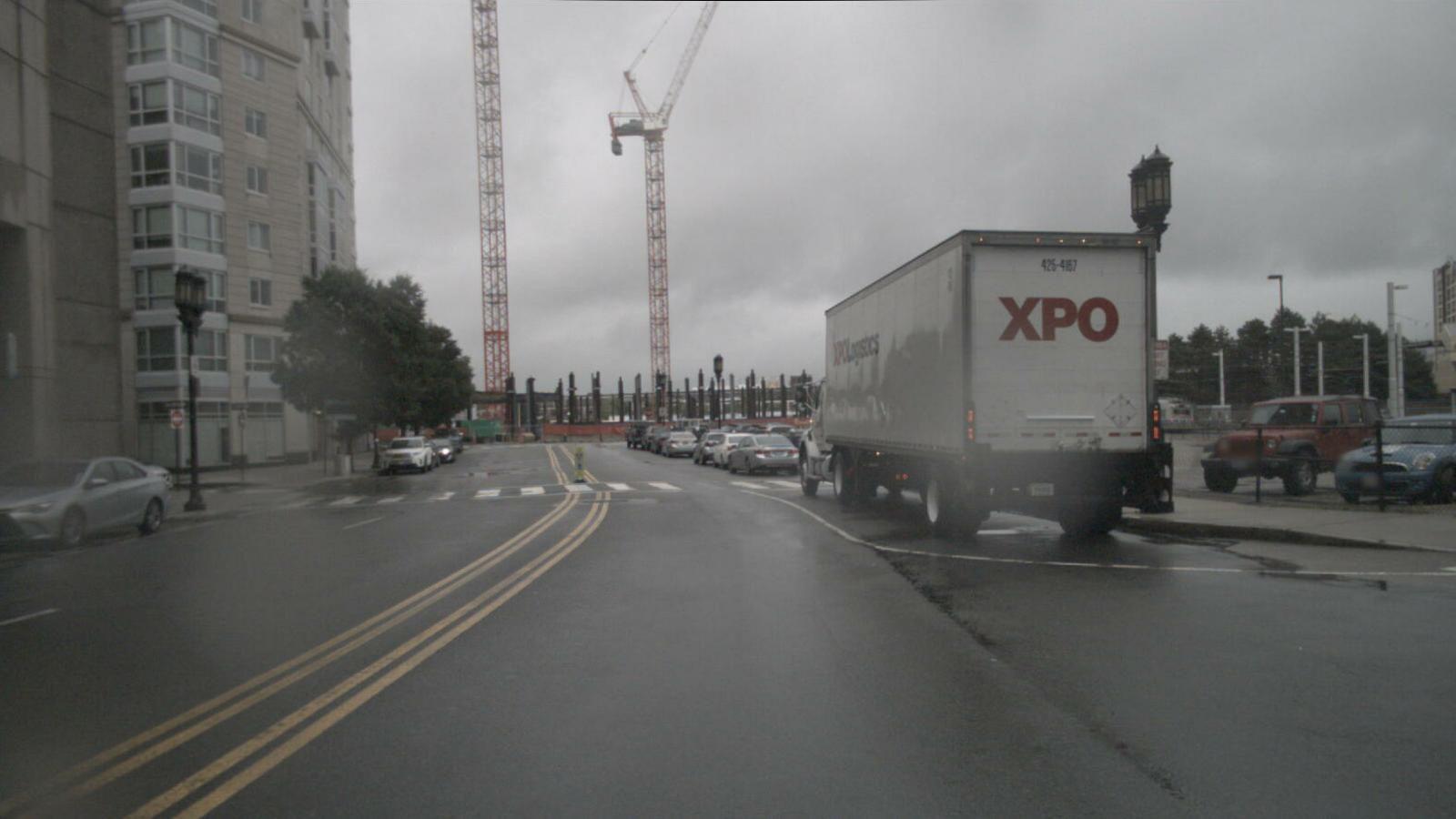} &
What is the future state of \texttt{<c3,\allowbreak CAM\_FRONT,\allowbreak 61.7,\allowbreak 548.3>}?&
A. Moving forward.\\
& & & B. Turning left into the intersection.\\
& & & C. Accelerating.\\
& & & \correct{D. Stationary.}\\[2pt]

\hypertarget{ex5}{}%
Planning &
\includegraphics[width=1.8cm,height=1.8cm,keepaspectratio]{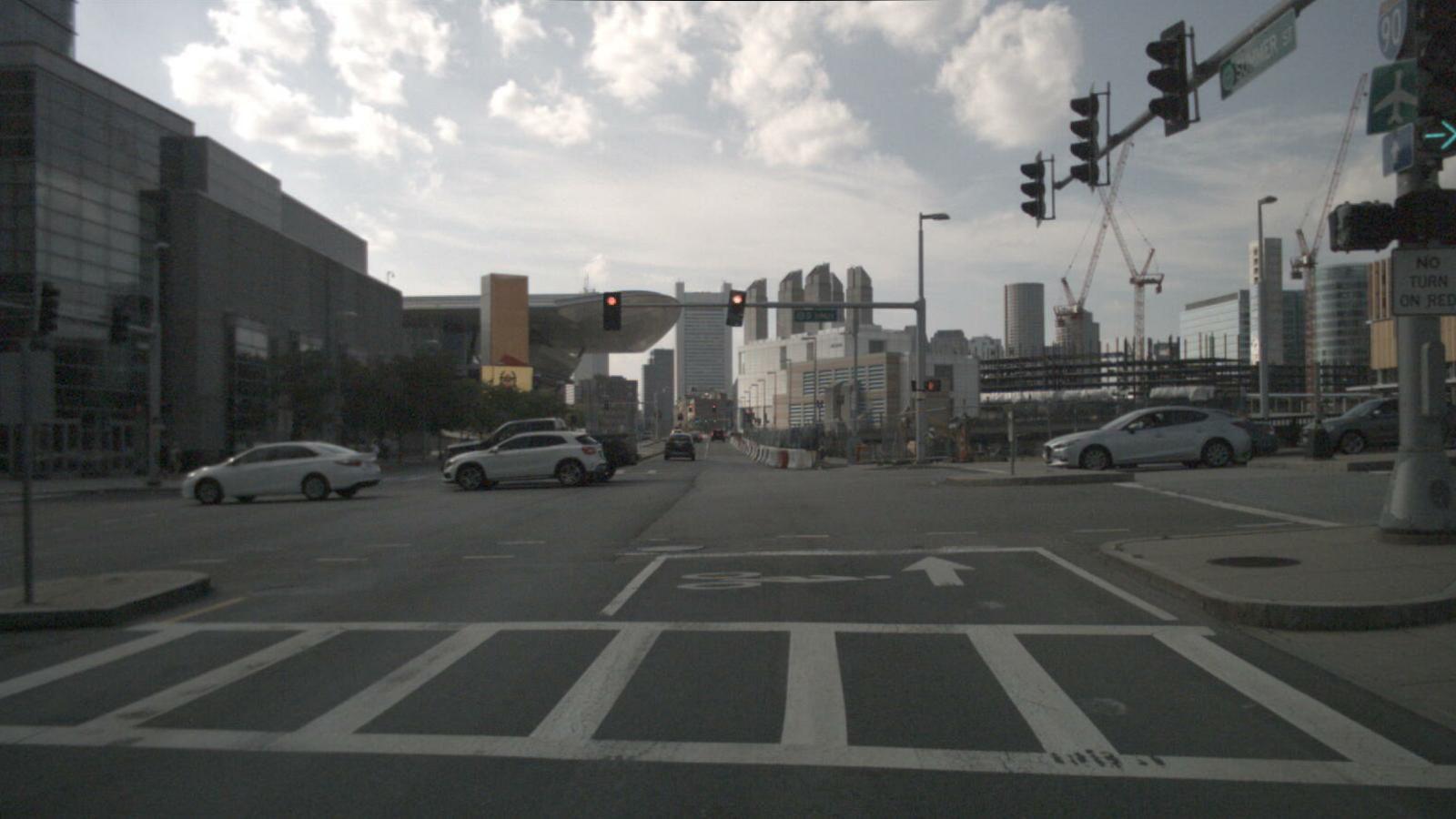} &
What actions could the ego vehicle take based on \texttt{<c3,\allowbreak CAM\_FRONT,\allowbreak 1298.3,\allowbreak 478.3>}?
Why take this action and what's the probability? &
A. Proceed forward through the intersection despite the red light.\\
& & & \correct{B. The action is to remain stationary. The reason for this action is to follow the traffic rules, with a high probability.}\\
& & & C. Change lanes into the left turn lane to proceed through the intersection.\\ 
& & & Turn right.\\[2pt]

\hypertarget{ex6}{}%
Planning &
\includegraphics[width=1.8cm,height=1.8cm,keepaspectratio]{\detokenize{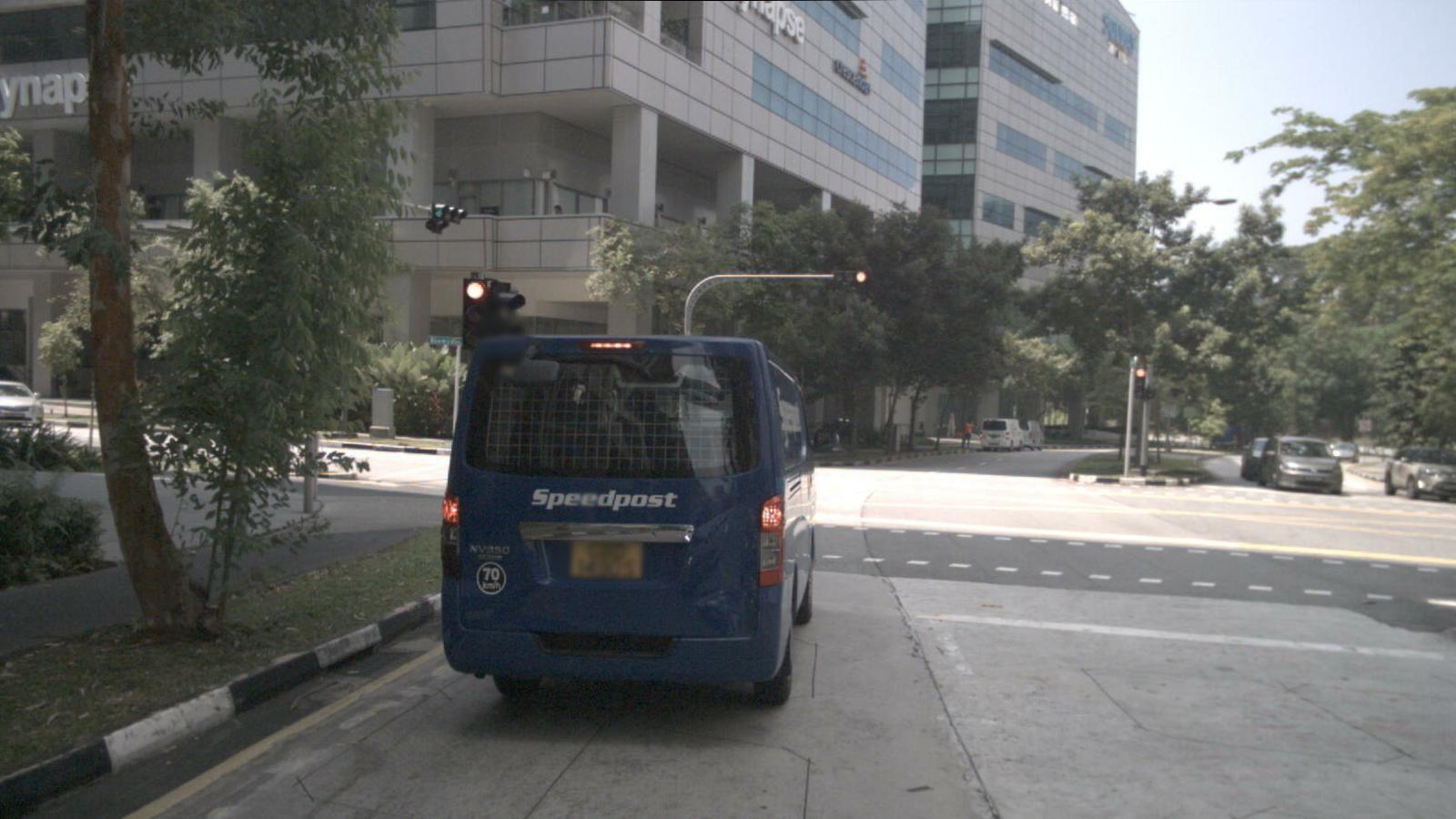}} &
Is \texttt{<c2,\allowbreak CAM\_FRONT,\allowbreak 708.7,\allowbreak 571.4>} an object that the ego vehicle should consider in the current scene? &
 \correct{A. Yes.}\\
& & & B. No, because stationary vehicles are irrelevant to decision-making.\\
& & & C. No, because the bus is outside the ego vehicle’s path.\\
& & & D. No, because only moving vehicles are considered relevant objects.\\

\end{xltabular}
\end{center}


\begin{table*}[t]
\centering
\caption{Per-example metrics for success cases from Table~\ref{tab:S_AutoDrive_examples}. The Ex. column refers back to the corresponding row. This table provides a qualitative comparison of metrics, showing questions and labels from our evaluation dataset along with exemplary predictions and their sample-level scores. Green indicates agreement with ground truth, while red indicates disagreement.}
\label{tab:S_AutoDrive_metrics}
\setlength{\tabcolsep}{6pt}
\renewcommand{\arraystretch}{1.15}
\begin{adjustbox}{max width=\textwidth}
\begin{tabular}{@{}c p{5cm} c c c c@{}}
\toprule
\textbf{Ex.} & \textbf{Prediction (model output)} & \textbf{BLEU} & \textbf{CIDEr} & \textbf{ChatGPT (/5)} & \textbf{AutoDrive(\%, ↑)} \\
\midrule
\hyperlink{ex1}{1} & \texttt{There is one traffic cone.}       & 10.19   & 21.30   & 4   & \textcolor{green}{A} \\
\hyperlink{ex2}{2} & \texttt{There is a car directly in front of the ego car.} & 11.17  & 13.92   & 2   & \textcolor{green}{C} \\
\hyperlink{ex3}{3} & \texttt{The traffic barrier.}                    & 0   & 0   & 0   & \textcolor{green}{C} \\
\hyperlink{ex4}{4} & \texttt{The future state of it is stationary.}              & 0.1   & 15.16   & 5   & \textcolor{green}{D}\\
\hyperlink{ex5}{5} & \texttt{The ego vehicle should remain stationary, as the traffic light ahead is red.}       & 0.28   & 7.68  & 4   & \textcolor{green}{B} \\
\hyperlink{ex6}{6} & \texttt{Yes, \texttt{<c2,\allowbreak CAM\_FRONT,\allowbreak 708.7,\allowbreak 571.4>} is an object that the ego vehicle should consider in the current scene.}                     & 0   & 0.096   & 5   & \textcolor{green}{A}\\
\bottomrule
\end{tabular}
\end{adjustbox}
\end{table*}
\end{OneColumnXLT}  
\vspace{0.75\baselineskip}
\begin{OneColumnXLT} 
\begin{center}
\begin{xltabular}{\textwidth}{@{}l c p{4cm} Y@{}}
\caption{Failure cases of AutoDrive-QA. These are examples where the model makes an incorrect judgment about the correctness of the prediction. Green indicates agreement with ground truth, while red indicates disagreement. Each example lists the task, image, question, and four choices (with the correct choice shown in \textcolor{orange}{orange}).}
\label{tab:AutoDrive_examples}\\
\toprule
\textbf{Task} & \textbf{Image} & \textbf{Question} & \textbf{Choices} \\
\midrule
\endfirsthead

\multicolumn{4}{c}{{\tablename\ \thetable} -- continued from previous page}\\
\toprule
\textbf{Task} & \textbf{Image} & \textbf{Question} & \textbf{Choices} \\
\midrule
\endhead

\midrule
\multicolumn{4}{r}{Continued on next page}\\
\endfoot

\bottomrule
\endlastfoot

\hypertarget{fx1}{}%
Perception &
\includegraphics[width=1.8cm,height=1.8cm,keepaspectratio]{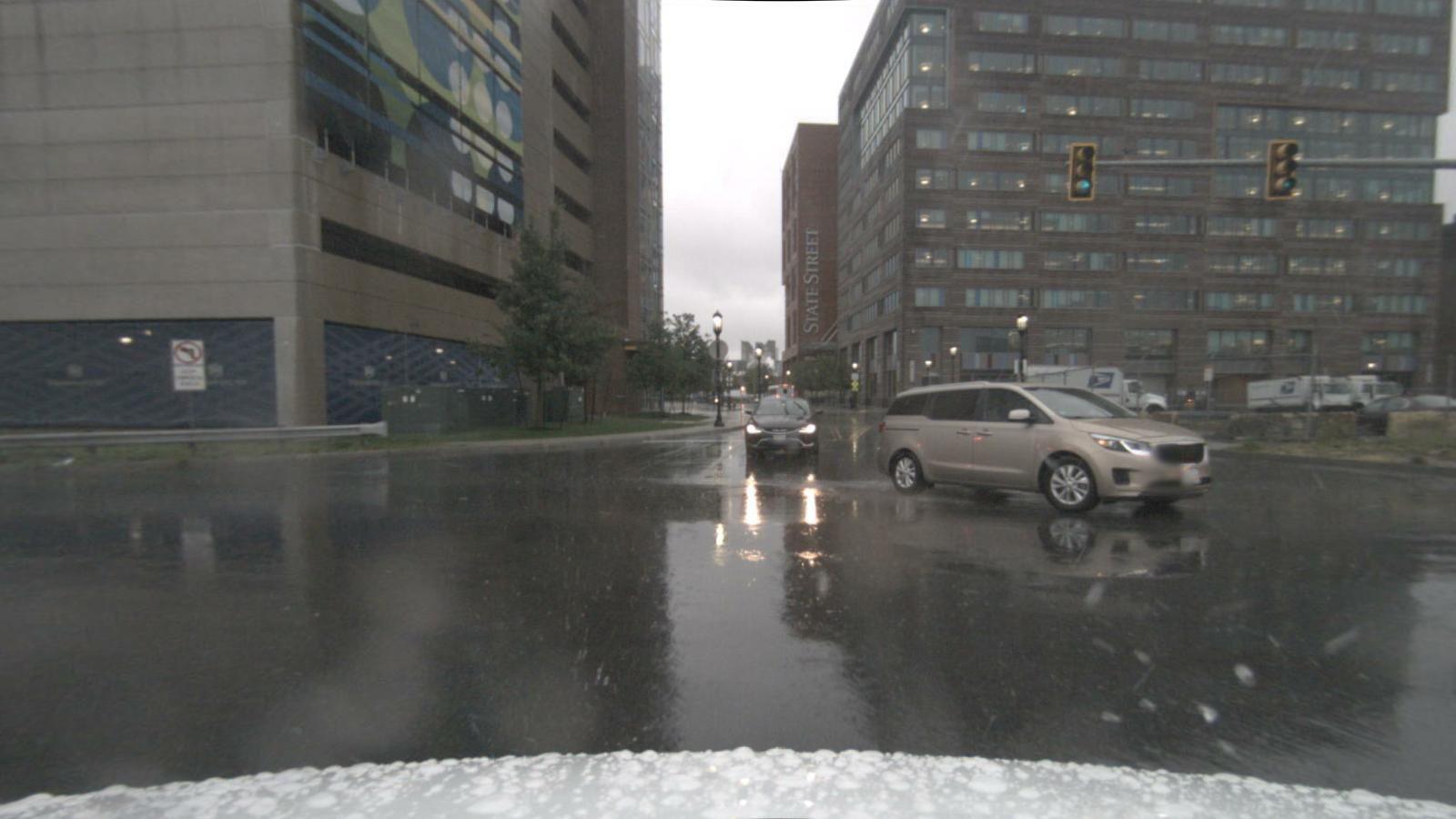} &
What is the status of the cars that are to the back of the ego car? &
\correct{A. Two cars are moving.}\\
& & & B. One car is moving, and the other is parked.\\
& & & C. Both cars are stationary.\\
& & & D. Two cars are parked.\\[2pt]

\hypertarget{fx2}{}%
Prediction &
\includegraphics[width=1.8cm,height=1.8cm,keepaspectratio]{n015-2018-09-25-11-10-38+0800__CAM_FRONT__1537845387662460.jpg} &
Based on the observations of \texttt{<c1,\allowbreak CAM\_BACK,\allowbreak 917.5,\allowbreak 687.5>}, what are possible actions to be taken by \texttt{<c2,\allowbreak CAM\_FRONT,\allowbreak 772.5,\allowbreak 512.5>}? What is the reason? &
A. The blue bus should accelerate, because the silver sedan is stationary.\\
& & & B. The blue bus should turn left.\\
& & & \correct{C. The action is none; the reason is that there is no safety issue.}\\
& & & D. The blue bus should stop immediately.\\[2pt]

\hypertarget{fx3}{}%
Planning &
\includegraphics[width=1.8cm,height=1.8cm,keepaspectratio]{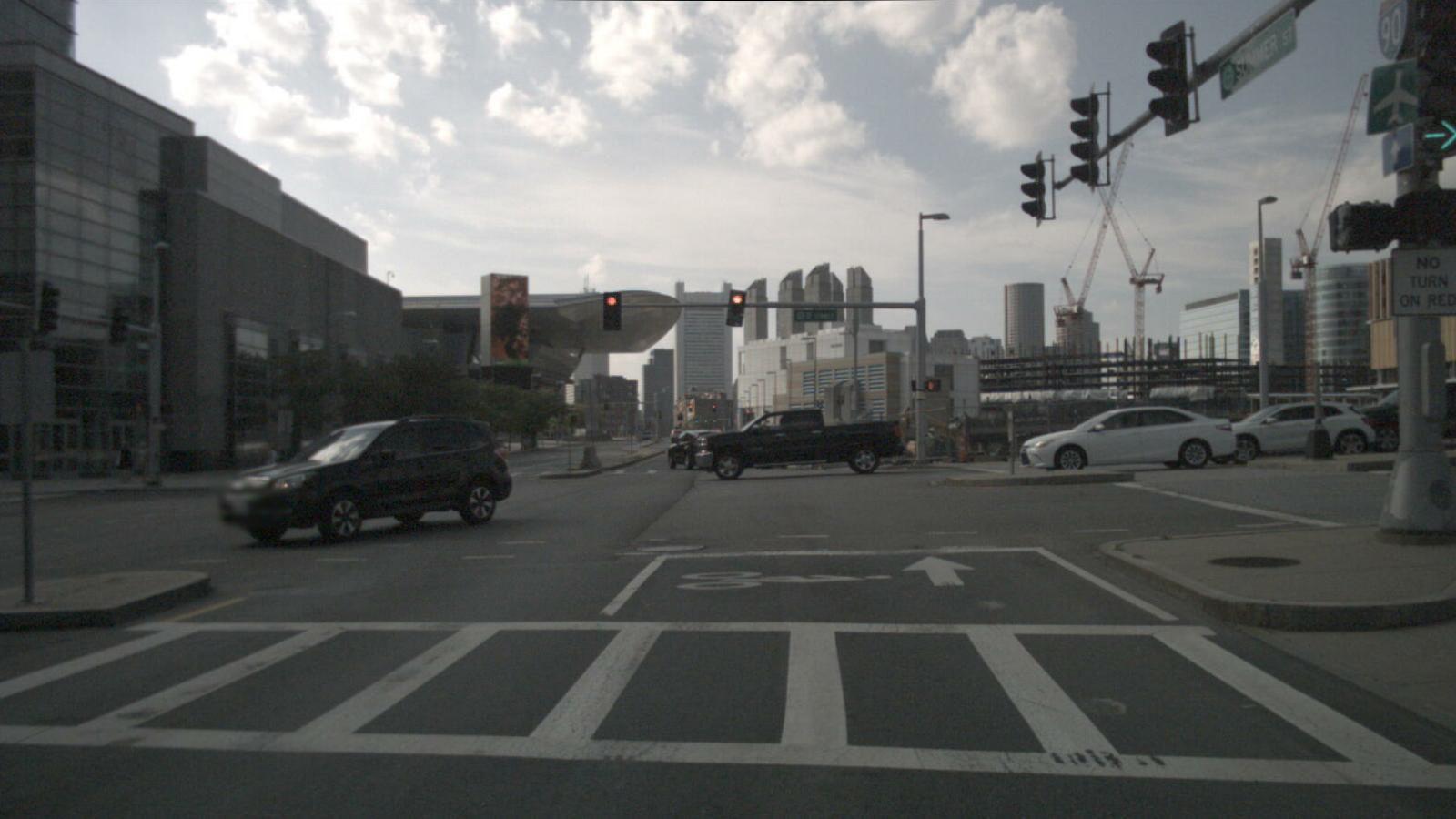} &
What is the priority of the objects that the ego vehicle should consider (in descending order)? &
A. The blue sedan.\\
& & & B. \texttt{<c1,\allowbreak CAM\_BACK\_RIGHT,\allowbreak 509.2,\allowbreak 633.3>},
\texttt{<c3,\allowbreak CAM\_BACK,\allowbreak 861.7,\allowbreak 675.0>}.\\
& & & \correct{C. The priority (descending) is:
\texttt{<c6,\allowbreak CAM\_FRONT,\allowbreak 734.3,\allowbreak 335.9>},
\texttt{<c2,\allowbreak CAM\_FRONT,\allowbreak 422.5,\allowbreak 519.2>},
\texttt{<c4,\allowbreak CAM\_FRONT,\allowbreak 878.3,\allowbreak 488.3>},
\texttt{<c5,\allowbreak CAM\_FRONT\_LEFT,\allowbreak 720.8,\allowbreak 586.7>},
\texttt{<c3,\allowbreak CAM\_BACK,\allowbreak 861.7,\allowbreak 675.0>},
\texttt{<c1,\allowbreak CAM\_BACK\_RIGHT,\allowbreak 509.2,\allowbreak 633.3>}.}\\
& & & D. The blue sedan \texttt{<c3>} and the traffic element \texttt{<c6>}.\\[2pt]

\end{xltabular}
\end{center}

\begin{table*}[t]
\centering
\caption{Per-example metrics for failure cases from Table~\ref{tab:AutoDrive_examples}. The Ex. column refers back to the corresponding row. This table provides a qualitative comparison of metrics, showing questions and labels from our evaluation dataset along with exemplary predictions and their sample-level scores. Green indicates agreement with ground truth, while red indicates disagreement.}
\label{tab:AutoDrive_metrics}
\setlength{\tabcolsep}{6pt}
\renewcommand{\arraystretch}{1.15}
\begin{adjustbox}{max width=\textwidth}
\begin{tabular}{@{}c p{5cm} c c c c@{}}
\toprule
\textbf{Ex.} & \textbf{Prediction (model output)} & \textbf{BLEU} & \textbf{CIDEr} & \textbf{ChatGPT(/5)} & \textbf{AutoDrive(\%, ↑)} \\
\midrule
\hyperlink{ex1}{1} & \texttt{The cars to the back of the ego car are stationary.}       & 0   & 8.95   & 0  & \textcolor{red}{B} \\
\hyperlink{ex2}{2} & The possible action for \texttt{<c2,\allowbreak CAM\_FRONT, \allowbreak 772.5,\allowbreak  512.5>} is to remain stationary. & 0   & 8.33   & 3  & \textcolor{red}{D} \\
\hyperlink{ex3}{3} & \texttt{first consider the traffic light. Next, it should pay attention to the vehicles}                    & 0  & 0   & 2   & \textcolor{red}{D} \\
\bottomrule
\end{tabular}
\end{adjustbox}
\end{table*}

\end{OneColumnXLT}  
\vspace{0.75\baselineskip}

\section*{Appendix D: Qualitative Evaluation Examples}
\addcontentsline{toc}{section}{Appendix D: Qualitative Evaluation Examples}
\label{Auto-drive evaluation Examples}

To further illustrate the evaluation framework of AutoDrive-QA, we present representative success and failure cases in Tables \ref{tab:S_AutoDrive_examples}-\ref{tab:AutoDrive_metrics}.
\subsection{Success Cases}
Table~\ref{tab:S_AutoDrive_examples} presents representative cases where the model successfully identified the ground-truth answers across perception, prediction, and planning tasks. These examples demonstrate that AutoDrive-QA produces clear and unambiguous questions: when models exhibit adequate scene understanding, they are able to consistently select the correct option. Table~\ref{tab:S_AutoDrive_metrics} reports the corresponding per-example metrics. Notably, we observe a systematic mismatch between traditional text-based metrics (e.g., BLEU, CIDEr) and correctness; even accurate predictions frequently receive low similarity scores. In contrast, the AutoDrive metric—multiple-choice accuracy—reliably aligns with task success, highlighting the limitations of free-text evaluation schemes. Overall, these findings indicate that AutoDrive accuracy offers a more consistent and task-relevant measure of model performance compared to conventional similarity-based metrics.

\subsection{Failure Cases}
Table \ref{tab:AutoDrive_examples} shows examples where the model selected an incorrect option. These highlight systematic weaknesses: in perception tasks, models may confuse object motion (e.g., mistaking parked vehicles as moving); in prediction tasks, they may fail to identify reasonable agent actions; and in planning, they may struggle with prioritizing relevant objects specially in ling correct answer. The distractors—designed to reflect realistic failure modes—are often deceptively plausible, making these errors especially diagnostic. Table \ref{tab:AutoDrive_metrics} reports the per-example metrics for these failures. Once again, we observe that traditional metrics assign partial credit to incorrect outputs, while AutoDrive’s MCQ accuracy clearly records the mistake.

\section*{Appendix E: Training Parameters}
\addcontentsline{toc}{section}{Appendix E: Training Parameters}
\label{Training Parameters}

All experiments were conducted on a Linux system equipped with three NVIDIA RTX A6000 GPUs (48 GB), using PyTorch~2.2.1 and CUDA~12.1. We fine-tuned \texttt{LLaVA-1.5-7B} on the entire dataset with LoRA, using \texttt{bf16} precision, a cosine learning-rate schedule with 3\% warm-up, and a learning rate of $2 \times 10^{-4}$ for LoRA. The weight decay was set to 0 for LoRA.

\section*{Appendix F: Distractor Analysis}
\addcontentsline{toc}{section}{Appendix F: Distractor Analysis}
\label{Distractor Analysis}
To better understand model behavior, we analyze the distribution of errors across different distractor types, allowing us to identify whether mistakes arise from visual misinterpretation, conceptual gaps, or reasoning failures. An ablation study with naive distractors further confirms the rigor of our pipeline: domain-specific distractors significantly reduce accuracy compared to naive alternatives, demonstrating that our curated distractors are not random noise but systematically expose model weaknesses. To capture these weaknesses more precisely, we categorize distractors into three groups: \textit{misinterpreted sensor inputs}, \textit{driving domain misconceptions}, and \textit{logical inconsistencies}. This categorization enables us to move beyond treating all wrong answers equally by revealing which type of distractor a model tends to select when it errs.  
Table~\ref{table:Error-Modeling-Performance} presents the error breakdown across perception, prediction, and planning tasks. In perception, most errors arise from driving domain misconceptions (45.74\%), such as misidentifying traffic signs or pedestrians, followed by logical inconsistencies (38.05\%) and misinterpreted sensor inputs (16.21\%). Prediction errors are also dominated by domain misconceptions (41.10\%), but misinterpreted sensor inputs play a larger role (36.99\%), reflecting challenges in interpreting dynamic cues required for forecasting. Logical inconsistencies account for 21.92\% of errors. In planning, however, logical inconsistencies dominate (48.86\%), highlighting critical reasoning flaws, while driving misconceptions remain substantial (35.23\%) and sensor misinterpretations are less common (15.91\%).  
Taken together, these results show that perception and prediction errors are primarily driven by difficulties in visual interpretation and sensor data, whereas planning is most critically affected by flawed reasoning and decision-making. This analysis highlights the need for vision--language models to integrate accurate visual context understanding with a strong conceptual grounding in traffic rules and driving etiquette. By systematically categorizing and tracking distractor selection, AutoDrive-QA provides one of the first detailed comparisons of error types across perception, prediction, and planning, offering actionable insights into the strengths and weaknesses of current models in urban driving scenarios.  

\begin{table*}[ht]
\centering
\caption{Distractor-Driven Error Analysis of VLM Models in Urban Scene Understanding: Comparing Diffrent Errors Across Driving Stages}
\label{table:Error-Modeling-Performance}
\resizebox{\linewidth}{!}{
\begin{tabular}{@{}lccc@{}}
\toprule
\textbf{Error Type}               & \textbf{Perception (\%)} & \textbf{Prediction (\%)} & \textbf{Planning (\%)} \\ \midrule
Driving Domain Misconceptions     & 45.74                    & 41.10                    & 35.23                  \\
Logical Inconsistencies           & 38.05                    & 21.92                    & 48.86                  \\
Misinterpreted Sensor Inputs      & 16.21                    & 36.99                    & 15.91                  \\ \bottomrule
\end{tabular}
}
\end{table*}

\section*{Appendix G: Ablation Studies on AutoDrive-QA}
\addcontentsline{toc}{section}{Appendix G: Ablation Studies on AutoDrive-QA}
\label{Ablation / Robustness Checks}

To evaluate the robustness of AutoDrive-QA, we perform ablation experiments that analyze the impact of different distractor design choices on model performance. Specifically, we compare realistic error-modeled distractors against more generic alternatives. In the first setup, we use a prompt without error modeling, which produces random and shallow distractors that are easy to eliminate. In the second setup, we apply error modeling, generating distractors that remain incorrect but are more plausible and contextually grounded.

As shown in Table~\ref{table:Error-Modeling}, the use of error modeling leads to noticeable accuracy drops for the Qwen2-VL-7B \cite{wang2024qwen2vl} model: about 7\% in perception, 9\% in prediction, and 7\% in planning. This drop highlights that the challenge comes not from the original questions themselves but from the stronger distractors. With naive distractors, models can easily rule out irrelevant options, often making the correct answer obvious. By contrast, error-modeled distractors require genuine reasoning about the scene, resulting in a more difficult and robust evaluation.

Correctness, also reported in Table~\ref{table:Error-Modeling}, serves as a key complementary metric. It improves with error-modeled distractors, ensuring that when open-ended questions are converted into multiple-choice format, exactly one option remains correct. Since the original question and ground-truth answer are preserved, the main challenge is to design distractors that are clearly incorrect yet still plausible. To validate this, we adopt a multi-agent system: each question is rated by GPT-4o on a 5-point scale, where a score of 5 indicates an unambiguous correct answer and a score of 1 indicates ambiguity.

\begin{table*}[t]
\centering
\caption{Impact of error-modeled distractors on accuracy and correctness across perception, prediction, and planning tasks}
\label{table:Error-Modeling}
\begin{adjustbox}{max width=\textwidth}
\begin{tabular}{@{}lcccccc@{}}
\toprule
\textbf{Error Type} & \multicolumn{2}{c}{\textbf{Perception}} & \multicolumn{2}{c}{\textbf{Prediction}} & \multicolumn{2}{c}{\textbf{Planning}} \\
\cmidrule(lr){2-3}\cmidrule(lr){4-5}\cmidrule(lr){6-7}
 & \textbf{Accuracy (↓) (\%)} & \textbf{Correctness (↑)} & \textbf{Accuracy (↓) (\%)} & \textbf{Correctness (↑)} & \textbf{Accuracy (↓) (\%)} & \textbf{Correctness (↑)} \\ 
\midrule
Naive      & 58.36 & 4.41 & 48.97 & 4.56 & 50.97 & 4.43 \\
Driving Domain Misconceptions    & 52.74 & 4.54 & 43.97 & 4.47 & 49.03 & 4.53 \\
Logical Inconsistencies  & 50.26 & 4.47 & 48.11 & 4.46 & 51.16 & 4.57 \\
Misinterpreted Sensor Inputs     & 56.60 & 4.45 & 41.02 & 4.42 & 59.74 & 4.44 \\
Computational Oversights  & 52.46 & 4.51 & 46.32 & 4.55 & 50.18 & 4.47 \\
Question Ambiguity       & 57.74 & 4.53 & 44.10 & 4.52 & 50.48 & 4.50 \\
All        & \textbf{51.64} & \textbf{4.57} & \textbf{39.74} & \textbf{4.60} & \textbf{43.50} & \textbf{4.55} \\
\bottomrule
\end{tabular}
\end{adjustbox}
\end{table*}


\section*{Appendix H: Distractor Categories and Prompt Designs}
\addcontentsline{toc}{section}{Appendix H: Distractor Categories and Prompt Designs}
\label{Distractor Categories and Prompt Designs}

To ensure diverse and challenging distractors, we designed five specialized prompt templates, each targeting a distinct error type in self-driving perception.

\begin{itemize}
    \item \textbf{Driving Domain Misconceptions Prompt.}  
    Generates distractors based on conceptual misunderstandings of driving scenes. The system creates plausible but incorrect options that hinge on fundamental misperceptions, such as confusing vehicle categories, misjudging whether an object is parked or moving, or mixing up left vs.\ right positions. These errors are illustrative of the kinds of mistakes autonomous perception systems (and human learners) often make when interpreting road environments.  

    \item \textbf{Logical Inconsistencies Prompt.}  
    Creates distractors that capture logical reasoning errors in driving contexts. Rather than misperceiving visual details, these distractors follow faulty inference chains—for example, drawing incorrect cause--effect relations, ignoring contextual signals, or overgeneralizing from a single observation. The goal is to simulate reasoning pitfalls (e.g., “a large truck must be stationary”) that test deeper understanding of driving logic.  

    \item \textbf{Misinterpreted Sensor Inputs Prompt.}  
    Targets errors in reading visual data from vehicle sensors (e.g., cameras). Distractors produced here reflect misinterpretations of spatial cues, shapes, or occluded objects—such as mistaking a pedestrian for a cone, or confusing the distance of a truck. These are designed to highlight the subtle challenges of interpreting complex, visually cluttered road environments.  

    \item \textbf{Computational Oversights Prompt.}  
    Generates distractors based on errors in handling sensor data and numerical inputs. Examples include swapped bounding box coordinates, incorrect velocity estimates, or misapplied unit conversions. These errors represent how simple mistakes in data processing pipelines (fusion, calibration, or scaling) can produce misleading outputs, making them an important category for realistic distractors.  

    \item \textbf{Question Ambiguity Prompt.}  
    Produces subtle, high-level distractors that exploit biases in how a question might be read or interpreted. For instance, a distractor may look correct if the respondent assumes all large vehicles are trucks. These options are especially challenging because they rely on misreadings of phrasing or context, forcing evaluators to distinguish between plausible but biased interpretations and the correct answer.  
\end{itemize}

The fusion prompt serves as a selection stage, curating the most effective distractors from across all five error categories. It balances relevance, difficulty, diversity, and subtlety, ensuring the final distractors form a coherent and challenging set. This step avoids redundancy and guarantees that distractors collectively test a wide range of perception and reasoning skills without overlapping.

\subsection{Example Prompt (Misinterpreted Sensor Inputs)}
To illustrate the structure of our prompt design, we provide a detailed template for generating distractors based on Misinterpreted Sensor Inputs. This template demonstrates how instructions are organized into sections (Given, Task, Output Format, and Remember), ensuring consistency and clarity across all prompt types. The complete version of the Misinterpreted Sensor Inputs prompt is shown in Figure \ref{fig:visual_prompt}.

This category targets errors in interpreting visual data from vehicle sensors (e.g., cameras). The generated distractors capture misreadings of spatial cues, shapes, or occluded objects—for instance, mistaking a pedestrian for a traffic cone or misjudging the distance of a truck. Such distractors highlight the subtle but critical challenges of interpreting complex and visually cluttered driving environments.
\begin{figure*}[t]
\centering
\begin{tcolorbox}[colback=gray!5, colframe=black!40, width=0.95\linewidth, sharp corners, boxrule=0.3pt]

Given:  
1. One or more images captured from the vehicle’s camera view.\\
2. An open-ended question referring to the image(s).\\
3. The correct answer to the question.\\

Your task:  
1. Carefully analyze the provided image(s) (for your own understanding only; do not output this).\\
2. Generate \{num\_choice\} plausible but incorrect distractor options that reflect visual interpretation errors. Each distractor should:  
   – Be directly connected to a misinterpretation of the image(s).\\
   – Appear reasonable at first glance.\\
   – Mislead due to subtle visual misunderstanding.\\
   – Contain a small but clear flaw in interpreting the visual information.\\
   – Differ in difficulty and type of misinterpretation.\\
3. Make sure you understand how the correct answer relates to specific visual features in the image(s).\\
4. Focus on common visual interpretation errors, including:  
   – Spatial Misinterpretation: Errors in distances, relative positions, lane placement, or perspective.\\
   – Color Confusion: Misinterpreting color-coded information, traffic signals, lane markings, or subtle color differences.\\
   – Detail Oversight: Overlooking small but important cues  pedestrians, cyclists, traffic signs, or lane markings.\\
   – Scale Misjudgment: Misinterpreting size, depth, or proximity of objects, vehicles, pedestrians, or cyclists.\\
   – Cross-Image Miscomparison: Incorrectly comparing or contrasting elements across multiple images or camera views.\\
5. Create a diverse set of distractors that test different aspects of visual interpretation and analysis.\\
6. Each distractor should represent a plausible but ultimately incorrect reading of the image(s).\\
7. Consider the type of image (e.g., photograph, diagram, street-view, aerial view, or multi-camera perspective) and design errors typical for that format.\\
8. Match the complexity of distractors to the difficulty of the question and answer.\\
9. When multiple images are provided, ensure some distractors exploit inconsistencies or relationships across them.\\
10. For each distractor, provide a short explanation (up to three sentences) describing why it seems plausible, the flaw it contains, and how it misleads interpretation.\\

Output format:  
– For each distractor, use the following format:  
\ \ \ \ Option:  
\ \ \ \ option: [Option text]  
\ \ \ \ reason: [Concise explanation (maximum 3 sentences) of why the distractor was created]  

Remember:  
– The goal is to design challenging but clearly incorrect options that highlight visual interpretation errors.\\
– Distractors should be realistic enough for learners to consider, but recognizably wrong with closer analysis.\\
– Focus exclusively on visual interpretation errors, not conceptual or reasoning mistakes.\\
– Distractors must directly reference the image(s), not just the general topic of the question.\\
– Keep them concise, consistent, and no more complex than the correct answer.\\
– Maintain uniform capitalization across all options, including the correct answer.\\
– For multiple images, consider errors that arise from comparing or misaligning visual information across them.\\
– Pay attention to relationships, cues, or differences that span multiple images, and design distractors that plausibly misinterpret these.\\

\end{tcolorbox}
\caption{Paraphrased prompt for generating distractors targeting \textit{visual interpretation errors}, including driving-specific cases (misinterpreted sensor inputs).}
\label{fig:visual_prompt}
\end{figure*}

\section{Filtering and Validation Prompt Designs}
\label{Filtering and Validation Prompt Designs}
We employ a four–stage refinement pipeline to ensure that generated distractors are 
challenging, diverse, and unambiguously incorrect and each stage has a distinct role.
\begin{itemize}
    \item \textbf{Reviewer Prompt.}  
    Designed to iteratively refine distractors generated for a specific error type (e.g., reasoning, visual, bias). For each distractor, the reviewer assesses its plausibility, alignment with the intended error type, and whether it could be mistakenly interpreted as correct. Effective distractors are retained, while others receive targeted suggestions to increase difficulty and deceptiveness without making them correct. This process prioritizes conceptual challenge over linguistic complexity. The full detailed prompt for \textit{Reviewer Prompt} is presented in Figure \ref{fig:review_prompt}.  
    
    \item \textbf{Refiner Prompt.}  
    Focuses on improving problematic distractors identified during evaluation. Using the correct answer and reviewer feedback as inputs, the refiner generates exactly three improved distractors that are plausible but clearly incorrect, aligned with misconceptions, and consistent in style. This ensures the final set includes one unambiguous correct answer and pedagogically useful distractors. The full detailed prompt for \textit{Refiner Prompt} is presented in Figure \ref{fig:refine_prompt}.    

    \item \textbf{Evaluator Prompt.}  
    Verifies the validity of a multiple-choice question by checking whether only one answer option can be considered correct. The marked correct answer is always treated as fixed, while distractors are examined for partial validity. A correctness score (1–5) is assigned, and feedback highlights problematic distractors. This guarantees conformity to the single-correctness principle and prevents ambiguity in assessment. The full detailed prompt for \textit{Evaluator Prompt} is presented in Figure \ref{fig:evaluator_prompt}. 

    \item \textbf{Selector Prompt.}  
    Guides the model to curate a subset of the strongest distractors from a larger pool across multiple error categories. The selection emphasizes image relevance, diversity, difficulty, subtlety, and educational value. By considering both distractor text and reasoning, the selector ensures the final set is challenging, non-repetitive, and complementary, producing a balanced and high-quality question set. The full detailed prompt for \textit{Selector Prompt} is presented in Figure \ref{fig:selector_prompt}. 
\end{itemize}

\begin{figure*}[t]
\centering
\begin{tcolorbox}[colback=gray!5, colframe=black!40, width=0.95\linewidth, sharp corners, boxrule=0.3pt]

Your role is to critically review and refine multiple-choice distractors that were generated under a specified \{type\} error category.  
The objective is to enhance their challenge and deceptiveness, while ensuring that each remains clearly incorrect in the context of driving perception tasks.\\

\textbf{Given:}\\
1. One or more images or sensor views from the ego vehicle’s perspective.\\
2. A question about the driving scene (e.g., objects, positions, or traffic elements).\\
3. The correct answer.\\
4. A set of distractor options tied to a specific error type (e.g., reasoning, visual, question bias).\\
5. The reasoning that explains why each distractor was initially created.\\

\textbf{Your Task:}\\
1. Assess each distractor’s effectiveness in testing comprehension while remaining definitively wrong.\\
2. Verify that the distractor aligns with the intended \{type\} error and fits the context of the driving scene.\\
3. Determine whether the distractor could be mistaken for the correct answer. If so, explain why and propose adjustments.\\
4. Endorse distractors that are sufficiently strong and challenging, only refining their clarity or style if needed.\\
5. For weaker distractors, provide concrete suggestions to raise their difficulty and deceptiveness  
   – without adding unnecessary length or modifiers, and without making them correct.\\
6. Ensure all evaluations and suggested revisions are concise, limited to four sentences or fewer.\\

\textbf{Guidelines:}\\
– Prioritize conceptual challenge and depth over superficial linguistic changes.\\
– If a distractor risks being valid, explicitly state this and recommend revisions to make it unambiguously wrong.\\
– Enhance plausibility by grounding refinements in realistic driving contexts (vehicles, pedestrians, signals, lane positions) while avoiding over-specificity.\\
– Provide improvements that make distractors more attractive or deceptive without undermining their incorrectness.\\
– Maintain a sharp and transparent distinction between distractors and the correct answer.\\

\textbf{Output Format:}\\
Option:\\
\ \ \ \ option: [Option text]\\
\ \ \ \ comment: [Evaluation of its effectiveness and suggested refinements]\\

\end{tcolorbox}
\caption{Reviewer prompt for analyzing and refining distractors in autonomous driving scenarios, ensuring they remain plausible yet incorrect while improving difficulty.}
\label{fig:review_prompt}
\end{figure*}

\begin{figure*}[t]
\centering
\begin{tcolorbox}[colback=gray!5, colframe=black!40, width=0.95\linewidth, sharp corners, boxrule=0.3pt]

You are an expert in assessment design and autonomous driving perception.\\
Your task is to refine multiple-choice distractors by applying general principles of question quality 
while addressing domain-specific misconceptions common in driving environments.\\

\textbf{Given:}\\
1. One or more images or sensor data from the ego vehicle’s surroundings.\\
2. An open-ended question about the driving scene.\\
3. The correct answer.\\
4. A set of distractor options related to driving errors, along with reviewer feedback.\\

\textbf{Your Task:}\\
1. Review reviewer feedback to assess whether the distractor is strong, weak, or unclear.\\
2. Check whether it reflects a plausible misconception in driving perception.\\
3. If effective, retain it and improve clarity or wording without changing its core flaw.\\
4. If weak, revise it to introduce a deeper or subtler conceptual mistake (e.g., misjudging occlusion, 
   inferring motion from a static view, or confusing sensor cues).\\
5. Ensure the distractor is plausible but clearly incorrect.\\

\textbf{Guidelines for refinement:}\\
-- The correct answer is fixed and cannot be changed.\\
-- Modify only distractors identified as problematic.\\
-- Keep strong distractors intact unless minor polishing is needed.\\
-- Maintain conciseness and stylistic consistency with the correct answer.\\
-- Focus on perception-based errors (category, direction, position, or status) but avoid making them valid.\\
-- Use reviewer feedback as a guide, but go beyond it when suggestions are incomplete.\\
-- Preserve the intended difficulty while prioritizing conceptual challenge over wording complexity.\\

\textbf{Output format:}\\
Option:\\
\ \ \ \ option: [Refined distractor text]\\
\ \ \ \ reason: [1--3 sentences explaining the conceptual error]\\

\textbf{Remember:}\\
-- Distractors must be believable but unambiguously wrong.\\
-- Avoid ambiguity that could make a distractor valid.\\
-- Keep them grounded in realistic driving contexts (vehicles, pedestrians, lanes, signals) without unnecessary detail.\\
-- Provide no commentary beyond the specified format.\\

\end{tcolorbox}
\caption{Unified refinement prompt for improving distractors in autonomous driving multiple-choice questions.}
\label{fig:refine_prompt}
\end{figure*}

\begin{figure*}[t]
\centering
\begin{tcolorbox}[colback=gray!5, colframe=black!40, width=0.95\linewidth, sharp corners, boxrule=0.3pt]

Your task is to evaluate a multiple-choice driving perception question (with accompanying images) and determine whether any distractor options could also be considered correct answers.  
The goal is to ensure that each question has exactly one valid answer, avoiding ambiguity in road-scene interpretation.\\

\textbf{Important:}\\
– The marked correct answer must always be accepted as valid, regardless of your own judgment.\\
– Never critique or re-evaluate the correct answer itself. Focus only on whether any distractors could also be interpreted as correct in the driving context.\\

\textbf{Scoring Scale:}\\
5 – Perfect: All distractors are clearly incorrect.\\
4 – Good: Distractors are mostly wrong but may contain minor elements of correctness.\\
3 – Fair: At least one distractor could be partially correct.\\
2 – Poor: At least one distractor could be equally correct.\\
1 – Invalid: Multiple distractors are just as valid as the marked correct answer.\\

\textbf{Provide:}\\
1. A score (1–5).\\
2. A short explanation highlighting any problematic distractors.\\
3. Suggested refinements for those distractors, if needed, to make them unambiguously incorrect.\\

\textbf{Remember:}\\
– Do not question the correctness of the labeled answer—it is always correct by definition.\\
– Focus exclusively on whether distractors could mislead by appearing valid in the driving scene (e.g., lane position, object status, or signals).\\

\end{tcolorbox}
\caption{Evaluator prompt for checking correctness of driving-scene distractors, ensuring only one valid answer is possible.}
\label{fig:evaluator_prompt}
\end{figure*}

\begin{figure*}[t]
\centering
\begin{tcolorbox}[colback=gray!5, colframe=black!40, width=0.95\linewidth, sharp corners, boxrule=0.3pt]

You are an expert Selection Agent tasked with curating the most challenging and high-quality distractor options for multiple-choice questions based on one or more provided images.  
Your goal is to select the best \{fusion\_selected\_choice\_num\} unique distractors from a pool of multiple options, ensuring a diverse, non-repetitive, and challenging set that is relevant to the given driving scene.\\

\textbf{Given:}\\
1. One or more images or sensor views from the ego vehicle’s perspective.\\
2. A dictionary of distractor options, grouped into five categories:  
   – Driving Domain Misconceptions (\{num\_choice\} options)\\
   – Logical Inconsistencies (\{num\_choice\} options)\\
   – Misinterpreted Sensor Inputs (\{num\_choice\} options)\\
   – Computational Oversights (\{num\_choice\} options)\\
   – Question Ambiguity (\{num\_choice\} options)\\
3. Each distractor is accompanied by a reason explaining its generation.\\

\textbf{Your Task:}\\
1. Carefully review all distractor options in relation to the driving context.\\
2. Select the top \{fusion\_selected\_choice\_num\} distractors using the following criteria:  
   – \textit{Image relevance}: Prioritize distractors closely tied to the road environment (vehicles, pedestrians, signals, lane positions).\\
   – \textit{Difficulty}: Choose options that require deeper understanding to recognize as incorrect.\\
   – \textit{Quality}: Favor distractors that are well-constructed and clearly tied to the correct answer.\\
   – \textit{Diversity}: Ensure coverage across different error categories.\\
   – \textit{Subtlety}: Prefer distractors with small but realistic mistakes (e.g., spatial or directional confusion).\\
   – \textit{Educational value}: Select options that, when revealed as wrong, provide insight into common driving misperceptions.\\
   – \textit{Uniqueness}: Ensure no two distractors overlap in meaning or phrasing.\\
   – \textit{Reason-based selection}: Factor in the provided reasoning, prioritizing those that align well with the driving scenario.\\

3. Guarantee balanced representation across categories, while allowing more selections from types highly relevant to the question.\\
4. Ensure the final set contains exactly \{fusion\_selected\_choice\_num\} distractors and excludes the correct answer.\\

\textbf{Output Format:}\\
Option:\\
\ \ \ \ option: [Option text]\\
\ \ \ \ reason: [Concise explanation, less than 3 sentences, of why it was selected]\\

\textbf{Remember:}\\
– The goal is to create a balanced set of distractors that test understanding of autonomous driving perception using the provided images.\\
– Ensure the chosen distractors work as a cohesive set, covering different aspects of perception and reasoning.\\
– Distractors must always be incorrect, concise, and stylistically consistent with the correct answer.\\
– Pay attention to visual and contextual cues in the driving scene when selecting distractors.\\
– If multiple images are provided, ensure selections remain valid across views or highlight inconsistencies between them.\\
– Avoid redundancy: do not include distractors that repeat the same idea in different wording.\\

\end{tcolorbox}
\caption{Selector prompt for curating the most challenging and diverse distractors in autonomous driving scenarios.}
\label{fig:selector_prompt}
\end{figure*}


\begin{thebibliography}{9}

\bibitem{chen2024}
Chen, L., Sinavski, O., Hünermann, J., Karnsund, A., Willmott, A.~J., Birch, D., and Shotton, J. (2024, May).
Driving with LLMs: Fusing object-level vector modality for explainable autonomous driving.
In \textit{Proceedings of the 2024 IEEE International Conference on Robotics and Automation (ICRA)}, 
pp. 14093--14100. IEEE.

\bibitem{nie2023}
Nie, M., Peng, R., Wang, C., Cai, X., Han, J., Xu, H., and Zhang, L. (2023).
Reason2Drive: Towards interpretable and chain-based reasoning for autonomous driving.
\textit{arXiv preprint arXiv:2312.03661}.

\bibitem{zhang2024}
Zhang, Y., Ma, Z., Li, J., Qiao, Y., Wang, Z., Chai, J., and Kordjamshidi, P. (2024).
Vision-and-language navigation today and tomorrow: A survey in the era of foundation models.
\textit{arXiv preprint arXiv:2407.07035}.

\bibitem{khalili2024}
Khalili, B. and Smyth, A.~W. (2024).
SOD-YOLOv8—enhancing YOLOv8 for small object detection in aerial imagery and traffic scenes.
\textit{Sensors}, 24(19):6209.


\bibitem{huang2024}
Huang, Z., Feng, C., Yan, F., Xiao, B., Jie, Z., Zhong, Y., and Ma, L. (2024).
DriveMM: All-in-one large multimodal model for autonomous driving.
\textit{arXiv preprint arXiv:2412.07689}.

\bibitem{sima2024}
Sima, C., Renz, K., Chitta, K., Chen, L., Zhang, H., Xie, C., and Li, H. (2024, September).
DriveLM: Driving with graph visual question answering.
In \textit{Proceedings of the European Conference on Computer Vision (ECCV)}, 
pp. 256--274. Cham: Springer Nature Switzerland.

\bibitem{chen2024eval}
Chen, K., Li, Y., Zhang, W., Liu, Y., Li, P., Gao, R., and Jia, X. (2024).
Automated evaluation of large vision-language models on self-driving corner cases.
\textit{arXiv preprint arXiv:2404.10595}.

\bibitem{marcu2024}
Marcu, A.~M., Chen, L., Hünermann, J., Karnsund, A., Hanotte, B., Chidananda, P., and Sinavski, O. (2024, September).
LingoQA: Visual question answering for autonomous driving.
In \textit{Proceedings of the European Conference on Computer Vision (ECCV)}, 
pp. 252--269. Cham: Springer Nature Switzerland.
\bibitem{wang2023drivemlm}
Wang, W., Xie, J., Hu, C., Zou, H., Fan, J., Tong, W., and Dai, J. (2023).
DriveMLM: Aligning multi-modal large language models with behavioral planning states for autonomous driving.
\textit{arXiv preprint arXiv:2312.09245}.

\bibitem{corbiere2025}
Corbière, C., Roburin, S., Montariol, S., Bosselut, A., and Alahi, A. (2025).
DRIVINGVQA: Analyzing visual chain-of-thought reasoning of vision-language models in real-world scenarios with driving theory tests.
\textit{arXiv preprint arXiv:2501.04671}.

\bibitem{mao2023}
Mao, J., Ye, J., Qian, Y., Pavone, M., and Wang, Y. (2023).
A language agent for autonomous driving.
\textit{arXiv preprint arXiv:2311.10813}.

\bibitem{tian2024}
Tian, X., Gu, J., Li, B., Liu, Y., Wang, Y., Zhao, Z., and Zhao, H. (2024).
DriveVLM: The convergence of autonomous driving and large vision-language models.
\textit{arXiv preprint arXiv:2402.12289}.
pp. 311--318.

\bibitem{fu2024}
Fu, D., Lei, W., Wen, L., Cai, P., Mao, S., Dou, M., and Qiao, Y. (2024, June).
Limsim++: A closed-loop platform for deploying multimodal LLMs in autonomous driving.
In \textit{Proceedings of the 2024 IEEE Intelligent Vehicles Symposium (IV)}, 
pp. 1084--1090. IEEE.

\bibitem{papineni2002}
Papineni, K., Roukos, S., Ward, T., and Zhu, W.~J. (2002, July).
BLEU: A method for automatic evaluation of machine translation.
In \textit{Proceedings of the 40th Annual Meeting of the Association for Computational Linguistics (ACL)}, 

\bibitem{banerjee2005}
Banerjee, S. and Lavie, A. (2005, June).
METEOR: An automatic metric for MT evaluation with improved correlation with human judgments.
In \textit{Proceedings of the ACL Workshop on Intrinsic and Extrinsic Evaluation Measures for Machine Translation and/or Summarization}, 
pp. 65--72.

\bibitem{lin2004}
Lin, C.~Y. (2004, July).
ROUGE: A package for automatic evaluation of summaries.
In \textit{Text Summarization Branches Out}, pp. 74--81.

\bibitem{vedantam2015}
Vedantam, R., Lawrence Zitnick, C., and Parikh, D. (2015).
CIDEr: Consensus-based image description evaluation.
In \textit{Proceedings of the IEEE Conference on Computer Vision and Pattern Recognition (CVPR)}, 
pp. 4566--4575.



\bibitem{zhang2025}
Zhang, Y., Su, Y., Liu, Y., Wang, X., Burgess, J., Sui, E., and Yeung Levy, S. (2025).
Automated generation of challenging multiple choice questions for vision-language model evaluation.
\textit{arXiv preprint arXiv:2501.03225}.

\bibitem{anderson2016}
Anderson, P., Fernando, B., Johnson, M., and Gould, S. (2016).
SPICE: Semantic propositional image caption evaluation.
In \textit{Computer Vision--ECCV 2016: 14th European Conference, Amsterdam, The Netherlands, October 11--14, 2016, Proceedings, Part V}, 
pp. 382--398. Springer International Publishing.

\bibitem{wen2023}
Wen, L., Yang, X., Fu, D., Wang, X., Cai, P., Li, X., Ma, T., Li, Y., Xu, L., Shang, D., Zhu, Z., Sun, S., Bai, Y., Cai, X., Dou, M., Hu, S., and Shi, B. (2023).
On the road with GPT-4V(ision): Early explorations of visual–language model on autonomous driving.  
\textit{arXiv preprint arXiv:2310.01412}.


\bibitem{wang2023}
Wang, W., Bao, H., Dong, L., Bjorck, J., Peng, Z., Liu, Q., Aggarwal, K., Mohammed, O.~K., Singhal, S., Som, S., and Wei, F. (2023).
Image as a foreign language: BEiT pretraining for vision and vision-language tasks.  
In \textit{Proceedings of the IEEE/CVF Conference on Computer Vision and Pattern Recognition (CVPR)}, pp. 19175--19186. IEEE.

\bibitem{jiang2023}
Jiang, A.~Q., Sablayrolles, A., Mensch, A., Bamford, C., Chaplot, D.~S., de las Casas, D., Bressand, F., Lengyel, G., Lample, G., Saulnier, L., Lavaud, L.~R., Lachaux, M.~A., Stock, P., Scao, T.~L., Lavril, T., Wang, T., Lacroix, T., and Sayed, W.~E. (2023).
Mistral 7B.  
\textit{arXiv preprint arXiv:2310.06825}.

\bibitem{he2021}
He, P., Gao, J., and Chen, W. (2021).
DeBERTaV3: Improving DeBERTa using ELECTRA-style pre-training with gradient-disentangled embedding sharing.
\textit{arXiv preprint arXiv:2111.09543}.

\bibitem{openai2023}
OpenAI. (2023).
GPT-4 technical report.
\textit{arXiv preprint arXiv:2303.08774}.

\bibitem{chib2023}
Chib, P.~S., and Singh, P. (2023).
Recent advancements in end-to-end autonomous driving using deep learning: A survey.  
\textit{arXiv preprint arXiv:2301.12345}.


\bibitem{caesar2020}
Caesar, H., Bankiti, V., Lang, A.~H., Vora, S., Liong, V.~E., Xu, Q., and Beijbom, O. (2020).
nuScenes: A multimodal dataset for autonomous driving.
In \textit{Proceedings of the IEEE/CVF Conference on Computer Vision and Pattern Recognition (CVPR)}, 
pp. 11621--11631.

\bibitem{dosovitskiy2017}
Dosovitskiy, A., Ros, G., Codevilla, F., Lopez, A., and Koltun, V. (2017, October).
CARLA: An open urban driving simulator.
In \textit{Proceedings of the Conference on Robot Learning (CoRL)}, 
pp. 1--16. PMLR.

\bibitem{cao2024}
Cao, X., Zhou, T., Ma, Y., Ye, W., Cui, C., Tang, K., and Zheng, C. (2024).
MapLM: A real-world large-scale vision-language benchmark for map and traffic scene understanding.
In \textit{Proceedings of the IEEE/CVF Conference on Computer Vision and Pattern Recognition (CVPR)}, 
pp. 21819--21830.

\bibitem{qian2024}
Qian, T., Chen, J., Zhuo, L., Jiao, Y., and Jiang, Y.~G. (2024, March).
NuScenes-QA: A multi-modal visual question answering benchmark for autonomous driving scenario.
In \textit{Proceedings of the AAAI Conference on Artificial Intelligence}, 
Vol. 38, No. 5, pp. 4542--4550.

\bibitem{wang2024}
Wang, S., Yu, Z., Jiang, X., Lan, S., Shi, M., Chang, N., and Alvarez, J.~M. (2024).
Omnidrive: A holistic LLM-agent framework for autonomous driving with 3D perception, reasoning and planning.
\textit{arXiv preprint arXiv:2405.01533}.

\bibitem{wang2024qwen2vl}
Wang, P., Bai, S., Tan, S., Wang, S., Fan, Z., Bai, J., and Lin, J. (2024).
Qwen2-VL: Enhancing vision-language model’s perception of the world at any resolution.
\textit{arXiv preprint arXiv:2409.12191}.

\bibitem{ding2024}
Ding, X., Han, J., Xu, H., Liang, X., Zhang, W., and Li, X. (2024).
Holistic autonomous driving understanding by bird’s-eye-view injected multi-modal large models.
In \textit{Proceedings of the IEEE/CVF Conference on Computer Vision and Pattern Recognition (CVPR)}, 
pp. 13668--13677.

\bibitem{yu2024}
Yu, H.~C., Shih, Y.~A., Law, K.~M., Hsieh, K.~Y., Cheng, Y.~C., Ho, H.~C., and Fan, Y.~C. (2024).
Enhancing distractor generation for multiple-choice questions with retrieval augmented pretraining and knowledge graph integration.
\textit{arXiv preprint arXiv:2406.13578}.


\bibitem{lu2022}
Lu, J., Ye, X., Ren, Y., and Yang, Y. (2022).
Good, better, best: Textual distractors generation for multiple-choice visual question answering via reinforcement learning.
In \textit{Proceedings of the IEEE/CVF Conference on Computer Vision and Pattern Recognition (CVPR)}, 
pp. 4921--4930.

\bibitem{luo2024}
Luo, H., Deng, Y., Shen, Y., Ng, S.~K., and Chua, T.~S. (2024).
Chain-of-exemplar: Enhancing distractor generation for multimodal educational question generation.
In \textit{Proceedings of the Annual Meeting of the Association for Computational Linguistics (ACL)}.


\end{thebibliography}
\end{document}